\documentclass[letterpaper, 10 pt, conference]{ieeeconf}
\IEEEoverridecommandlockouts
\usepackage{graphicx} 
\usepackage{epsfig} 
\usepackage{times} 
\usepackage{amsmath} 
\usepackage{amssymb}  
\usepackage{amsfonts}
\usepackage{xcolor}
\usepackage{setspace}
\usepackage{hyperref}
\usepackage{subcaption}
\usepackage{algorithm}
\usepackage{algorithmic}

\usepackage{booktabs} 
\usepackage{multirow}
\usepackage{float}

\newlength{\defaulttextfloatsep}
\newlength{\defaultintextsep}
\begin{document}

\title{Learning Domain Randomization Distributions for Training Robust Locomotion Policies}

\author{Melissa Mozifian\textsuperscript{*}$^{1}$, Juan Camilo Gamboa Higuera\textsuperscript{*}$^{1}$, David Meger$^{1}$ and Gregory Dudek$^{1}$
\thanks{*Equal contribution.  $^{1}$ Montreal Institute of Learning Algorithms (MILA), and the Mobile Robotics Lab (MRL) at the School of Computer Science, McGill University, Montreal, Canada. Correspondence to: Melissa Mozifian {\tt\small melissa.mozifian@mcgill.ca}, Juan Camilo Gamboa Higuera {\tt\small gamboa@cim.mcgill.ca}}%
}

\maketitle

\begin{abstract}
Domain randomization (DR) is a successful technique for learning robust policies for robot systems, when the dynamics of the target robot system are unknown. The success of policies trained with domain randomization however, is highly dependent on the correct selection of the randomization distribution. The majority of success stories typically use real world data in order to carefully select the DR distribution, or incorporate real world trajectories to better estimate appropriate randomization distributions.
In this paper, we consider the problem of finding good domain randomization parameters for simulation, without prior access to data from the target system. We explore the use of gradient-based search methods to learn a domain randomization with the following properties: 1) The trained policy should be successful in environments sampled from the domain randomization distribution 2) The domain randomization distribution should be wide enough so that the experience similar to the target robot system is observed during training, while addressing the practicality of training finite capacity models.
These two properties aim to ensure the trajectories encountered in the target system are close to those observed during training, as existing methods in machine learning are better suited for interpolation than extrapolation. We show how adapting the domain randomization distribution while training context-conditioned policies results in improvements on jump-start and asymptotic performance when transferring a learned policy to the target environment. 
\end{abstract}


\section{Introduction}
\label{Introduction}
Machine learning, and deep reinforcement learning (deep-RL) in particular, is a promising
approach for learning controllers or action policies for complex systems where analytic methods are elusive.
The use of robot simulators for machine learning is driven by a compromise between the large data requirements of Deep-RL and the cost of collecting experience on physical robots systems. Except for simple robot systems in controlled environments, however, real robot experience may not correspond to situations that can be simulated; an issue known as the \emph{reality gap} ~\cite{jakobi1995noise}. This gap limits the utility of simulation-based learning.

\emph{Domain randomization} (DR) aims to address the reality gap by training policies to maximize performance over a diverse set of simulation models, where the parameters of each model are sampled randomly from a given task distribution. The effectiveness of DR has been demonstrated by training controllers purely in simulation that subsequently transfer successfully to real robot systems~\cite{peng2018sim, openai2018learning} and by fine-tuning the DR distributions with real world data ~\cite{chen2018hardware}.

\begin{figure}[t!]
\centering
    \begin{subfigure}[b]{0.21\textwidth}
        \includegraphics[width=\textwidth]{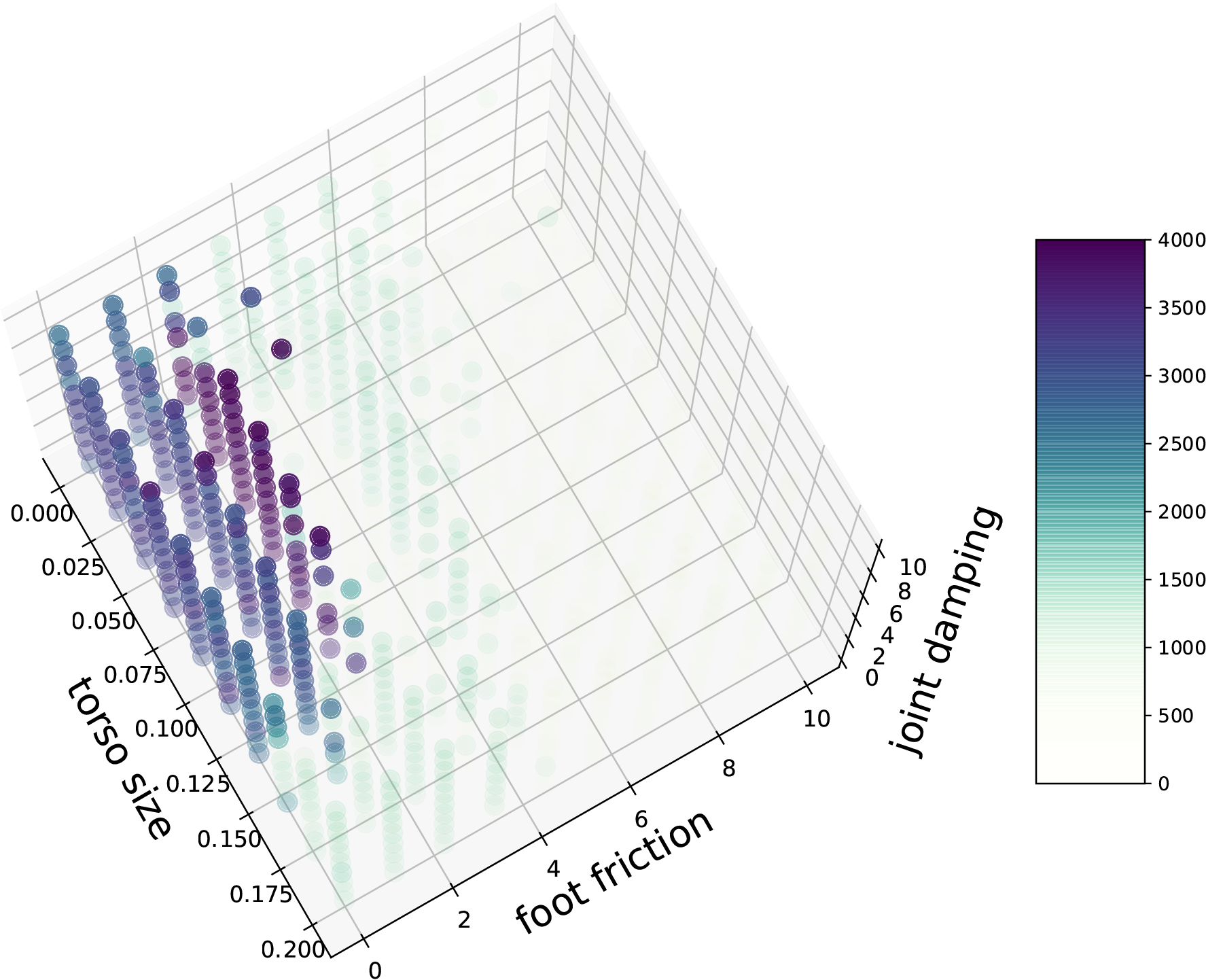}
        \caption{\small Average rewards with independently trained policies per context}
        \label{fig:hopper_target_rewards}
    \end{subfigure}
    ~
    \begin{subfigure}[b]{0.21\textwidth}
        \includegraphics[width=\textwidth]{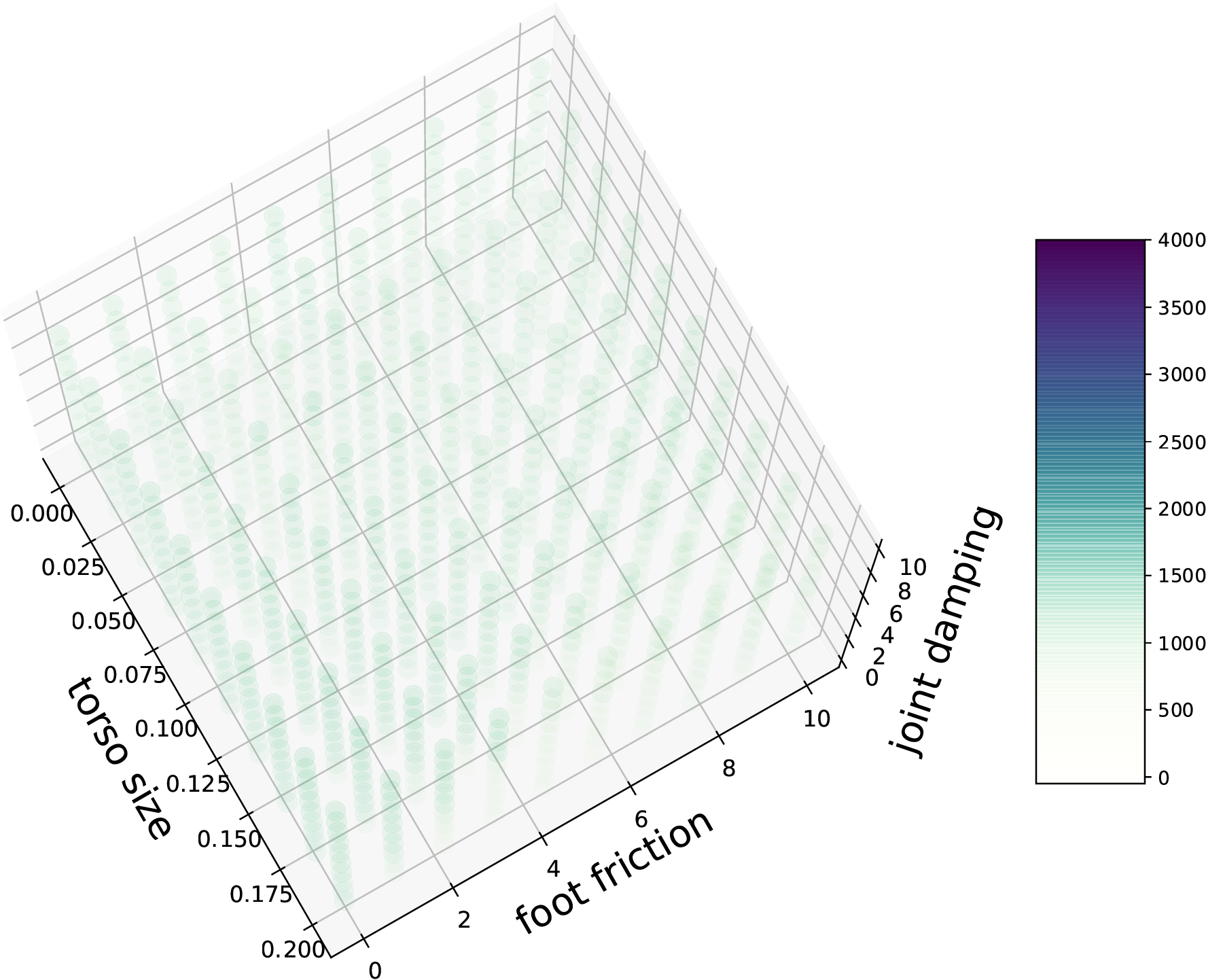}
        \caption{\small Average rewards with fixed, uniform domain randomization}
        \label{fig:hopper_fixed_dr}
    \end{subfigure}
    \begin{subfigure}[b]{0.21\textwidth}
        \includegraphics[width=\textwidth]{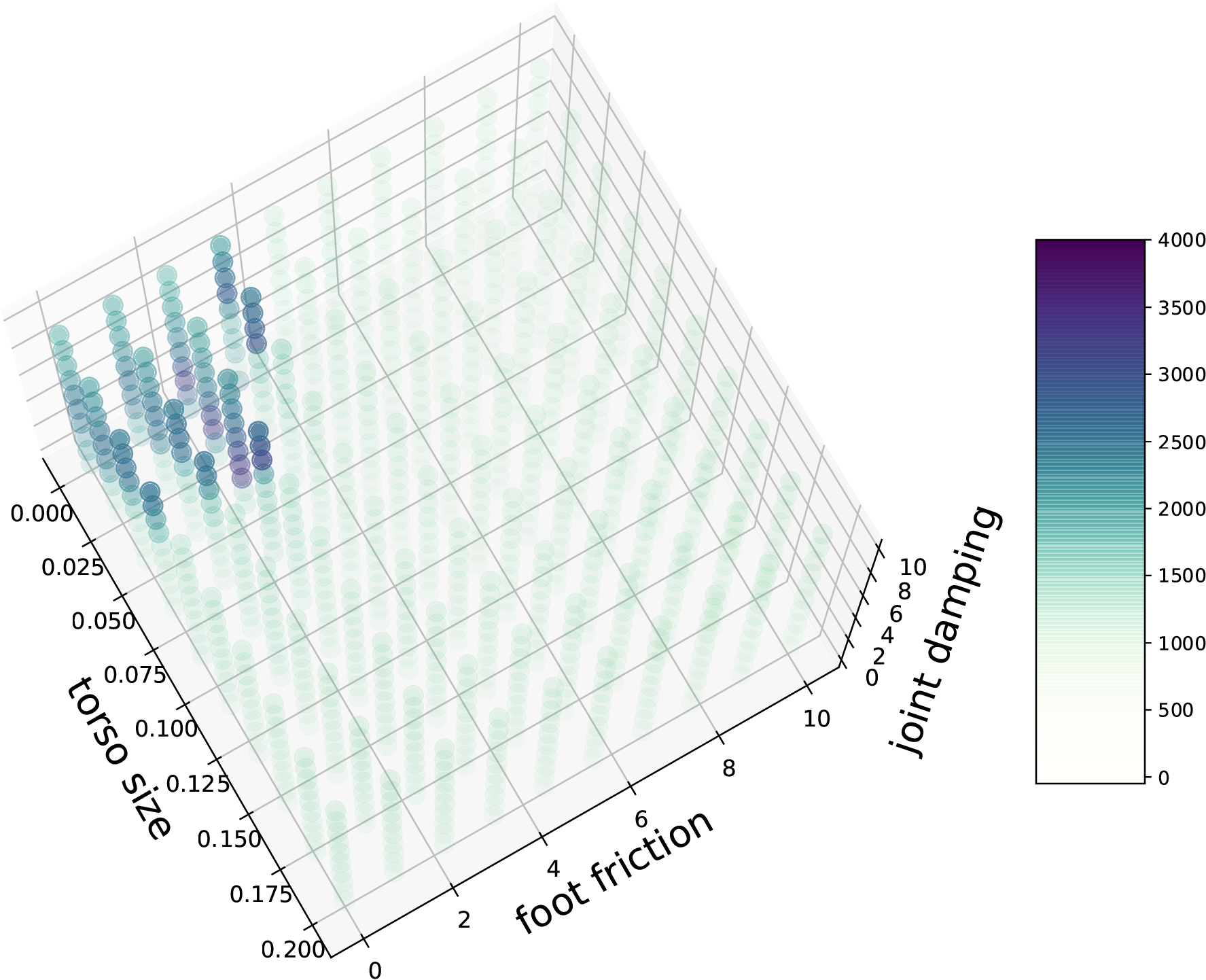}
        \caption{\small Average rewards with learned Gaussian domain randomization}
        \label{fig:hopper_learned_dr}
    \end{subfigure}
    ~
    \begin{subfigure}[b]{0.21\textwidth}
        \includegraphics[width=\textwidth]{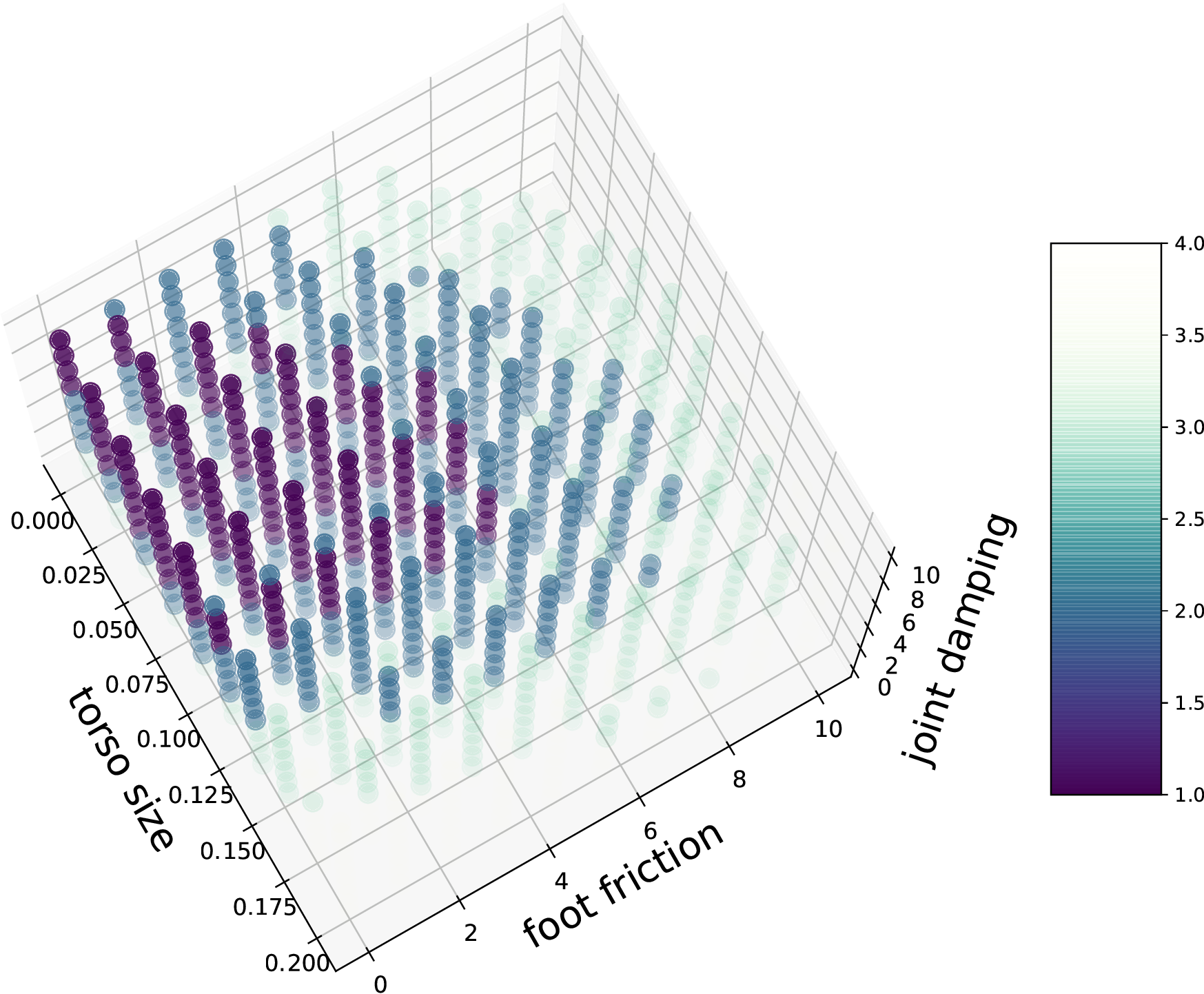}
        \caption{\small Confidence regions of the learned distribution for 1: 68\%, 2: 95\%, and 3: 99\% confidence}
        \label{fig:hopper_confidence}
    \end{subfigure}
    \caption{\small Figure~\ref{fig:hopper_target_rewards} shows the maximum rewards obtained by exhaustively running PPO on a grid of simulation parameters, ~\ref{fig:hopper_fixed_dr} shows the performance of a policy trained with uniform domain randomization over the grid, ~\ref{fig:hopper_learned_dr} results from our method when learning the parameters of a Gaussian domain randomization distribution (whose confidence regions are shown in ~\ref{fig:hopper_confidence}). For fixed and learned DR the reward averages were obtained by evaluating the policy with 10 different rollouts.}
    \label{fig:hopper_grid_comparison}
\end{figure}

While successful, an aspect that has not been addressed in depth is the selection of the domain randomization distribution. 
The DR distribution should be selected carefully to ensure the ability to generalize within the simulated experience. Selecting a distribution that is too wide may result in an under-performing policy (see Fig.~\ref{fig:hopper_fixed_dr}) or slow convergence (see Fig.~\ref{fig:epopt_minmax}) . Conversely, training  on  a distribution  that  is  too  narrow  leads  to  policies  that  don't generalize well~\cite{rajeswaran2016epopt,packer2018assessing}. Most of the existing methods~\cite{rajeswaran2016epopt, chebotar2018closing, ramos2019bayessim}, assume the availability of real robot data. In this work, we study the extreme case where no real robot data is available, similar to~\cite{openai2018learning}, in which the appropriate selection of a DR distribution is crucial.

We propose to learn the parameters of the simulator distribution, such that the policy is trained under an evolving sequence of simulators that gives priority to environments where the current policy can achieve high performance. Our proposed method aims to find a compromise between variety and performance, by optimizing the DR distribution to focus on environments where high rewards are likely, while regularizing the distribution parameters via a KL-divergence objective (explained in Sec.~\ref{sec:method}) and using more data from the worst performing simulators (see the experiments in Sec~\ref{sec:epopt}). Our algorithm learns the domain randomization distribution while simultaneously optimizing the policy to maximize performance over the learned distribution. Figure~\ref{fig:hopper_learned_dr} shows our resuls towards finding the \emph{sweet-spot} distribution.
We evaluate our method on a variety of control problems from the OpenAI Gym suite of benchmarks~\cite{gym2016}. We find that our method is able to improve on the performance of fixed domain randomization. Furthermore, we demonstrate our model's robustness to initial simulator distribution parameters, showing that our method repeatably converges to similar domain randomization distributions across different experiments.

\section{Related Work}

Developing robust or generalizable controllers has a long and rich history beyond the scope of this paper~\cite{Zames1981,zhou1998robustControl}. 
Caruana was on of the first to deeply explore the advantages of multi-task learning in a robotics context~\cite{Caruana93multitasklearning}. Baxter considered multi-task learning in a Bayesian/information-theoretic context~\cite{Baxter1997}.
%
%

Packer et al~\cite{packer2018assessing} present an empirical study of generalization in Deep-RL, testing interpolation and extrapolation performance of state-of-the-art algorithms when varying simulation parameters in control tasks. The authors provide an experimental assessment of generalization under varying training and testing distributions. Our work extends these results by providing results for the case when the training distribution parameters are learned and change during policy training.

Chebotar et al~\cite{chebotar2018closing} propose training policies on a distribution of simulators, whose parameters are fit to real-world data. Their proposed algorithm switches back and forth between optimizing the policy under the DR distribution and updating the DR distribution by minimizing the discrepancy between simulated and real world trajectories. In contrast, we aim to learn policies that maximize performance over a diverse distribution of environments where the task is feasible, as a way of minimizing the interactions with the real robot system. Similarly, Ramos et al~\cite{ramos2019bayessim} adopt a likelihood-free inference method that computes the posterior distribution of simulation parameter using real-world trajectories.  Rajeswaran et al~\cite{rajeswaran2016epopt} propose a related approach for learning robust policies over a distribution of simulator models. The proposed approach, based on the the $\epsilon$-percentile conditional value at risk (CVaR) ~\cite{tamar2015optimizing} objective, improves the policy performance on a small proportion of environments where the policy performs the worst. The authors propose to update the distribution of simulation models to maximize the likelihood of real-world trajectories, via Bayesian inference.
In Section~\ref{sec:result} we combine our method with the CVaR objective to encourage diversity of the learned DR distribution.

Related to learning the DR distribution, Paul et al~\cite{fpo_opt} propose using Bayesian Optimization (BO) to update from the simulation model distribution. This is done by evaluating the improvement over the current policy by using a policy gradient algorithm with data sampled from the current simulator distribution. The parameters of the simulator distribution for the next iteration are selected to maximize said improvement. Mehta et al~\cite{mehta2019adr} direct the policy search towards harder instances given the domain randomization ranges.  
Given a set reference environment, the domain randomization distribution is modified in order to produce simulated environments within which the learned policy behaves differently than the original policy. Their method explores the parameter space by learning a discriminative reward computed from discrepancies in policy roll-outs, generated in randomized and reference environments.

Other related methods rely on policies that are conditioned on \emph{context}: variables used to represent unobserved physical properties that can be varied in the simulator, either explicitly or implicitly. For example, Chen et al~\cite{chen2018hardware} propose learning a policy conditioned on the hardware properties of the robot, encoded as a vector $v_h$. These represent variations on the dynamics of the environment (i.e. friction, mass), drawn from a fixed simulator distribution. When explicit, the context is equal to the simulator parameters. When implicit, the mapping between context vectors and simulator environments is learned during training, using policy optimization. At test time, when the true context is unknown, the context vector $v_h$ that is fed as input to the policy is obtained by gradient descent on the task performance objective. Similarly, Yu et al~\cite{strategy_optim} propose training policies conditioned on simulator parameters (explicitly), then optimizing the context vector alone to maximize performance at test time. Training is done by collecting trajectories on a fixed simulation model distribution. The argument of the authors is that searching over context vectors is easier than searching over policy parameters. The proposed method relies on population-based gradient-free search for optimizing the context vector to maximize task performance. Our method follows a similar approach, but we focus on learning the training distribution. Rakelly et al~\cite{rakelly2019efficient} also use context-conditioned policies, where the context is implicitly encoded into a vector $z$. During the training phase, their proposed algorithm improves on the performance of the policy while learning a probabilistic mapping from trajectory data to context vectors. At test time, the learned mapping is used for online inference of the context vector. This is similar in spirit to the Universal Policies with Online System Identification method~\cite{yu2017preparing}, which instead uses deterministic context inference with an explicit context encoding. Again, these methods use a fixed DR distribution and could benefit from adapting it during training, as we propose in this work.

\section{Problem Statement}
We consider parametric Markov Decision Processes (MDPs) ~\cite{sutton2018reinforcement}.
An MDP $\mathcal{M}$ is defined by the tuple 
$\langle \mathcal{S},\mathcal{A}, p, r,\gamma, \rho_0\rangle$, where $\mathcal{S}$ is the set of possible states and $\mathcal{A}$ is the set of actions, $p$ : $\mathcal{S} \times \mathcal{A} \times \mathcal{S} \xrightarrow{} \mathbb{R}$, encodes the state transition dynamics, $r$ : 
$\mathcal{S} \times \mathcal{A} \xrightarrow{} \mathbb{R}$
is the task-dependent reward function, $\gamma$ is a discount factor, and
$\rho_0$ : $\mathcal{S} \xrightarrow{} \mathbb{R} $ is the initial state distribution.
Let $s_t$ and $a_t$ be the state and action taken at time $t$. At the beginning of each episode, $s_0 \sim \rho_0(.)$. Trajectories $\tau$ are obtained by iteratively sampling actions using the current policy, $\pi$, $a_t \sim \pi (a_t | s_t)$ and evaluating next states according to the transition dynamics $s_{t+1} \sim p(s_{t+1} | s_t, a_t, z)$, where $z$ are the parameters of the dynamics.
Given an MDP $\mathcal{M}$, the goal is then to learn policy $\pi$ to maximize the expected sum of rewards $J_{\mathcal{M}} (\pi) = \mathbb{E}_{\tau}\left[R(\tau) | \pi \right] = \mathbb{E}_{\tau}\left[\sum_{t=0}^\infty \gamma^t r_t\right]$, where $r_t=r(s_t,a_t)$.

In our work, we aim to maximize performance over a \emph{distribution} of MDPs, each described by a \emph{context} vector $z$ representing the variables that change over the distribution: changes in transition dynamics, rewards, initial state distribution, etc. Thus, our objective is to maximize $\mathbb{E}_{z \sim p_(z)}\left[J_{\mathcal{M}_{z}} (\pi)\right]$, where $p(z)$ is the domain randomization distribution. Similar to~\cite{strategy_optim,chen2018hardware, rakelly2019efficient}, we condition the policy on the context vector, $\pi({a_t|s_t, z})$. In the experiments reported in this paper, we let $z$ encode the parameters of the transition model in a physically based simulator; e.g. mass, friction or damping.

\section{Proposed Method}
\label{sec:method}
We introduce \textbf{LSDR} (Learning the Sweet-spot Distribution Range) algorithm for concurrently learning a domain randomization distribution and a robust policy that maximizes performance over it. LSDR requires a prior distribution $p(z)$, which is a guess of what the domain randomization ranges should be. Instead of directly sampling from $p(z)$, we use a surrogate distribution $p_{\phi}(z)$, with trainable parameters $\phi$. Our goal is to find appropriate parameters $\phi$ to optimize $\pi(\cdot| s, z)$. LSDR proceeds by updating the policy with trajectories sampled from $p_{\phi}(z)$, and updating the $\phi$ based on the performance of the policy $p(z)$. To avoid the collapse of the learned distribution, we use a regularizer that encourages the distribution to be diverse. The idea is to sample more data from environments where improvement of the policy is possible, without collapsing to environments that are trivial to solve.
We summarize our training and testing procedure in Algorithm~\ref{alg:LSDR-policy} and ~\ref{alg:LSDR-distr}.
In our experiments, we use Proximal Policy Optimization (PPO)~\cite{schulman2017proximal} for the $\textrm{UpdatePolicy}$ procedure in Algorithm~(\ref{alg:LSDR-policy}).

\begin{algorithm}[th]
   \caption{\small Learning the policy and training distribution}
   \label{alg:LSDR-policy}
\begin{algorithmic}
   \REQUIRE testing distribution $p(z)$, initial parameters of the learned distribution $\phi$, initial policy $\pi$, buffer size $B$, total iterations $N$
    \FOR {$i \in \{1,...,N\}$}
    \STATE     $z \sim p_{\phi}(z)$
    \STATE     $s_0 \sim \rho_0(s)$
    \STATE     $\mathcal{B} = \{\}$
    \WHILE{    $|\mathcal{B}| < B$}
    \STATE         $a_{t} \sim \pi(a_{t}| s_{t}, z)$
    \STATE         $s_{t+1}, r_{t} \sim p(s_{t+1}, r_{t} | s_{t}, a_{t}, z)$
    \STATE         append $(s_{t}, z_k, a_{t}, r_{t}, s_{t+1})$ to $\mathcal{B}$
    \IF{         $s'$ is terminal}
    \STATE             $z \sim p_{\phi}(z)$
    \STATE             $s_{t+1} \sim \rho_0(s)$
    \ENDIF
    \STATE       $s_t \leftarrow s_{t+1}$
    \ENDWHILE
    \STATE     $\phi \leftarrow \textrm{UpdateDistribution}(\phi, p(z), \pi)$
    \STATE     $\pi \leftarrow \textrm{UpdatePolicy}(\pi, \mathcal{B})$
    \ENDFOR
\end{algorithmic}
\end{algorithm}

\begin{algorithm}[th]
   \caption{\small UpdateDistribution}
   \label{alg:LSDR-distr}
\begin{algorithmic}
   \REQUIRE learned distribution parameters $\phi$, testing distribution $p(z)$, policy $\pi$, total iterations $M$, total trajectory samples $K$
    \FOR {$i \in \{1,...,M\}$}
    \STATE sample $z_{1:K}$ from $p(z)$
    \STATE Obtain Monte-Carlo estimate of $\mathcal{L}_{DR}(\phi)$ by executing $\pi$ on environments with $z_{1:K}$
    \STATE $\phi \leftarrow \phi + \lambda \nabla_{\phi} (\mathcal{L}_{DR}(\phi) - \alpha D_{KL}(p(z)||p_{\phi}(z)))$
    \ENDFOR
\end{algorithmic}
\end{algorithm}

\subsection{Learning the Sweet-spot Distribution Range}
The goal of our method is to find a training distribution $p_{\phi}(z)$ to maximize the expected reward of the policy over $p(z)$, while gradually reducing the sampling frequency of environments where the task is not solvable\footnote{In this work, we consider a task solvable if there exists a policy that brings the environment to a set of desired goal states.}. Such a situation is common in physics-based simulations, where a suboptimal selection of simulation parameters may lead to environments where the task is impossible due to physical limits or unstable simulations.

We start by assuming that the prior distribution $p(z)$ has wide but bounded support, such that we get a distribution of solvable and unsolvable tasks. To update the training distribution, we use an objective of the following form
\begin{equation}
    \underset{\phi}{\arg\max\,}\mathcal{L}_{DR}(\phi) - \\ \alpha D_{KL}(p(z)||p_{\phi}(z))
    \label{eq:dr_objective}
\end{equation}
where the first term is designed to encourage improvement on environments that are more likely to be solvable, while the second term is a regularizer that keeps the distribution from collapsing. In our experiments, we set $\mathcal{L}_{DR}(\phi) = \mathbb{E}_{z \sim p(z)} [J_{\mathcal{M}_{z}} (\pi)\log(p_{\phi}(z))]$ so that the distribution is driven towards high reward regions.
If we use the performance of the policy as a way of determining whether the task is solvable for a given context $z$, then a trivial solution would be to make $p_{\phi}(z)$ concentrate on few easy environments. The second term in Eq.~(\ref{eq:dr_objective}) helps to avoid this issue by penalizing distributions that deviate too much from $p(z)$, which is assumed to be wide.

To estimate the gradient of Eq.~(\ref{eq:dr_objective}) with respect to $\phi$, we use the log-derivative score function gradient estimator~\cite{fu2006gradient}, resulting in the following Monte-Carlo update :

\begin{equation}
\begin{split}
\phi \leftarrow \phi + \lambda \bigg[ \frac{1}{K} \sum_{i=1}^K \bigg( J_{\mathcal{M}_{z_i}}(\pi) \nabla_{\phi}\log(p_{\phi}(z_i))  \bigg)  - \\
\alpha \nabla_{\phi}D_{KL}(p(z)||p_{\phi}(z))  \bigg]
\end{split}
\end{equation}

where $z_{i} \sim p(z)$. Updating $\phi$ with samples from the distribution we are learning, can be problematic as we may not get information about the performance of the policy in low probability contexts under $p_{\phi}(z)$. 

\begin{figure}[t!]
\centering
    \begin{subfigure}[b]{0.2\textwidth}
        \includegraphics[width=\textwidth]{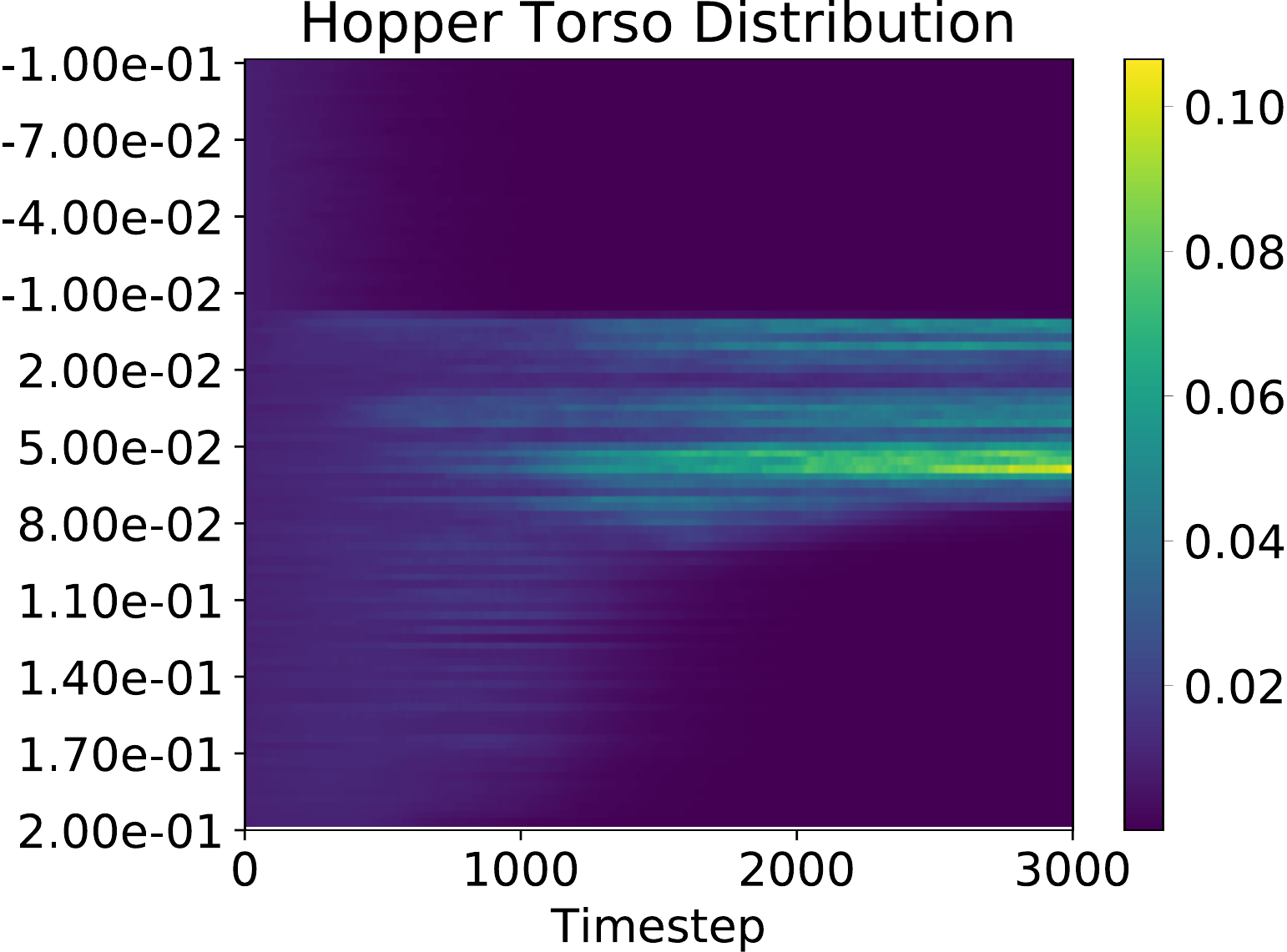}
        \caption{}
        \label{fig:no_entropy}
    \end{subfigure}
    ~
    \begin{subfigure}[b]{0.2\textwidth}
        \includegraphics[width=\textwidth]{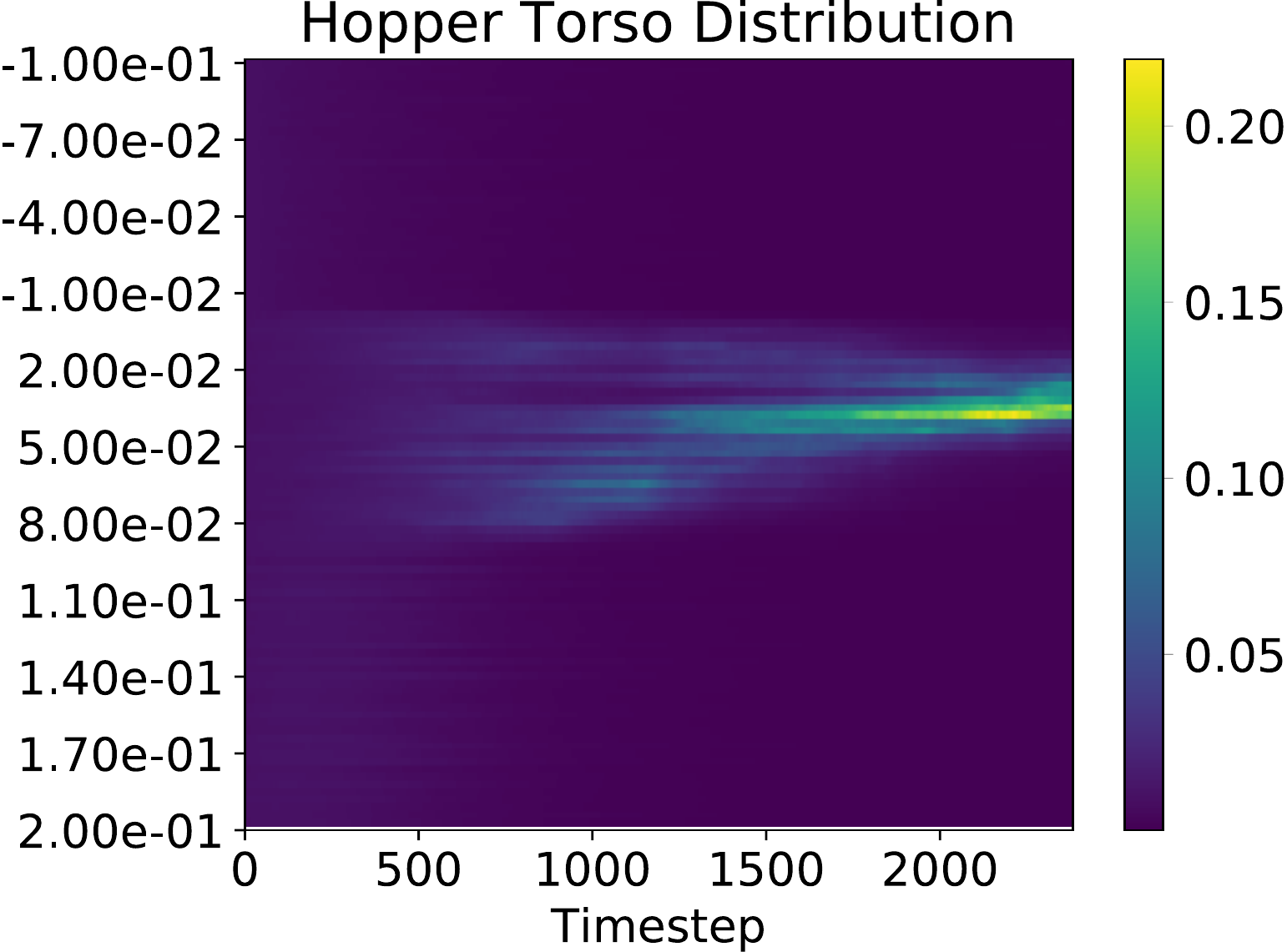}
        \caption{}
        \label{fig:train_sampl}
    \end{subfigure}
    \caption{\small Learned torso size distribution for Hopper. Figure ~\ref{fig:no_entropy} shows distribution learned without the KL divergence regularizer and Figure~\ref{fig:train_sampl} shows the distribution learned while sampling from train distribution.}
    \label{fig:not-learning}
\end{figure}

Empirically, we observed this issue in our single dimensional experiments with discrete distributions, where the distribution collapses as it never sees low-probability samples (see Figure~\ref{fig:not-learning}). An option in that case would be to use samples from $p(z)$ to evaluate the first term of our objective, which biases the gradients but avoids the distribution collapse. For the experiments with multivariate Gaussian distributions changing the sampling distribution was not required.

To ensure that the two terms in Eq.~(\ref{eq:dr_objective}) have similar scale, we standardize the evaluations of $J_{\mathcal{M}_{z_i}}$ with exponentially averaged batch statistics and set $\alpha$ via hyper-parameter search. 

\section{Experimental Results}
\label{sec:result}
\begin{figure}[t!]
    \centering
    \begin{subfigure}[b]{0.15\textwidth}
        \includegraphics[width=\textwidth]{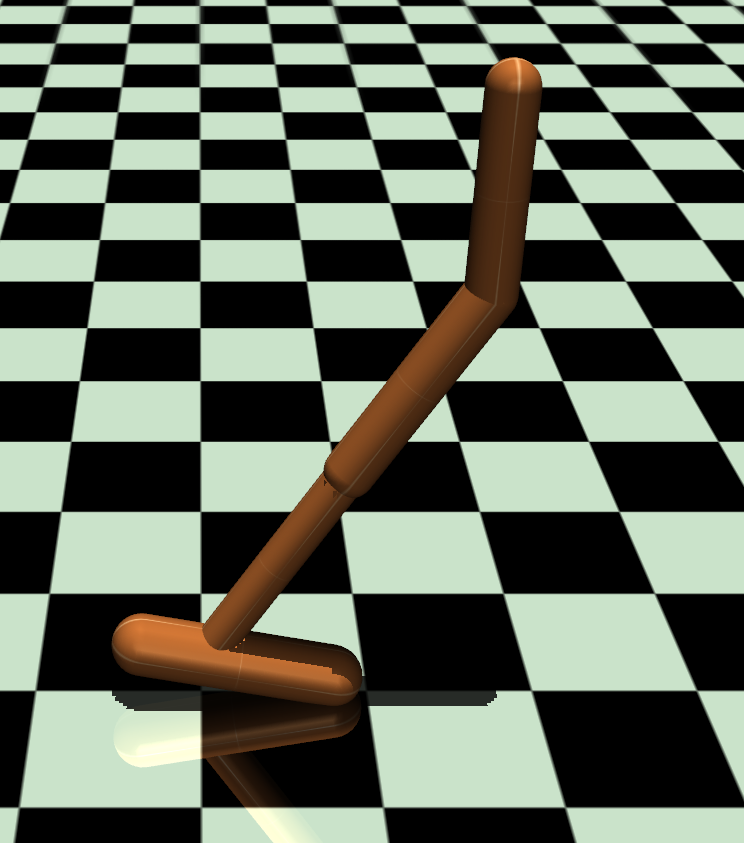}
        \caption{}
        \label{fig:hophop}
    \end{subfigure}
    ~
    \begin{subfigure}[b]{0.23\textwidth}
        \includegraphics[width=\textwidth]{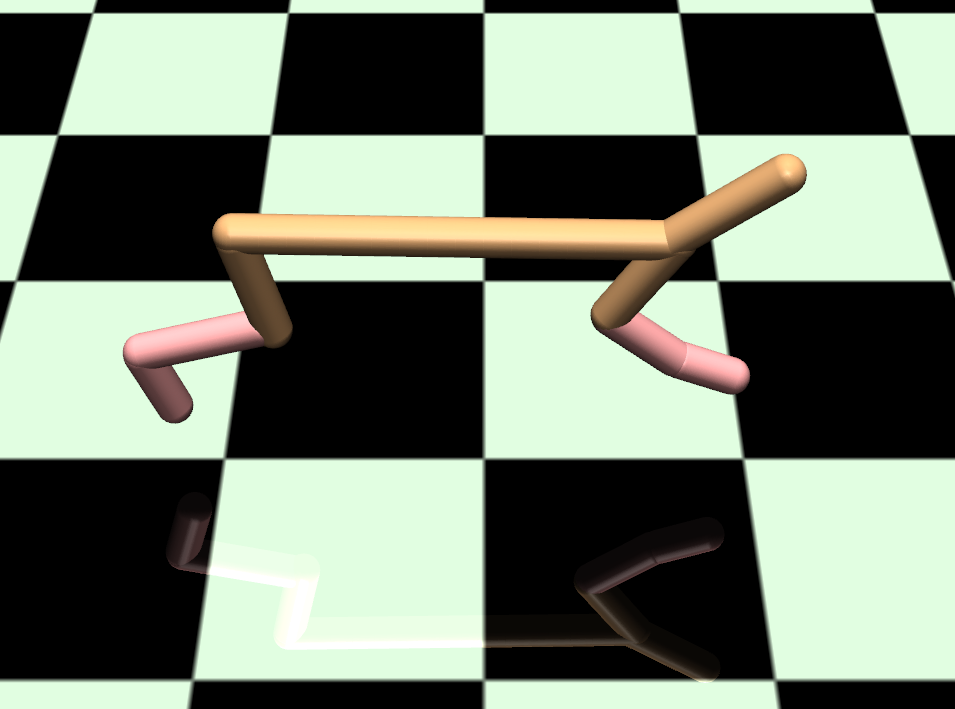}
        \caption{}
        \label{fig:halfcheetah}
    \end{subfigure}
    \caption{\small Illustrations of the 2D simulated robot models used in the experiments. The hopper (a) and half-cheetah (b) tasks, present more challenging environments when varying dynamics.}\label{fig:mujoco-viz}
\end{figure}

We evaluate the impact of learning the DR distribution on two standard benchmark locomotion tasks: Hopper and Half-cheetah, see Fig.~\ref{fig:mujoco-viz}, from the MuJoCo tasks in the OpenAI Gym suite~\cite{gym2016}. We use an explicit encoding of the context vector $z$, corresponding to the torso size, density, foot friction and joint damping of the environments. Our experiments study both single-dimensional as well as multidimensional domain randomization contexts. For the first, we use discrete distributions and run experiments for each context dimension separately. For latter, we parametrize the domain randomization distribution with multivariate Gaussians.

We selected the prior $p(z)$ as an uniform distribution over ranges that include both solvable and unsolvable environments. In the case of discrete distributions, we discretize the support of $p(z)$ with 100 bins, and initialize $p_{\phi}(z)$ to be uniform; i.e. equivalent to $p(z)$. When sampling from this distribution, we first select a bin according to the discrete probabilities, then select a continuous context value uniformly at random from the ranges of the corresponding bin. For the multidimensional context experiments, the mean of the Gaussian distribution is initialized with the mean of test range, and the variance is initialized to $1/10^{th}$ of the variance of $p(z)$, so that most of the mass of the distribution falls within its support.

We compare the test-time jump-start and asymptotic performance of policies learned with $p_{\phi}(z)$ (learned domain randomization) and $p(z)$ (fixed domain randomization). At test time, we sample (uniformly at random) a test set of samples from the support of $p(z)$, $50$ for single-dimensional and $100$ for multidimensional contexts, and fine-tune the policy with the parameters obtained at training time. The questions we aim to answer with our experiments are: 1) does learning policies with wide DR distributions affect the performance of the policy in the environments where the task is solvable? 2) does learning the DR distribution converge to the actual ranges where the task is solvable? 3) Is learning the DR distribution beneficial?

\subsection{Results}
\begin{figure}[t!]
    \centering
    \begin{subfigure}[b]{0.2\textwidth}
        \includegraphics[width=\textwidth]{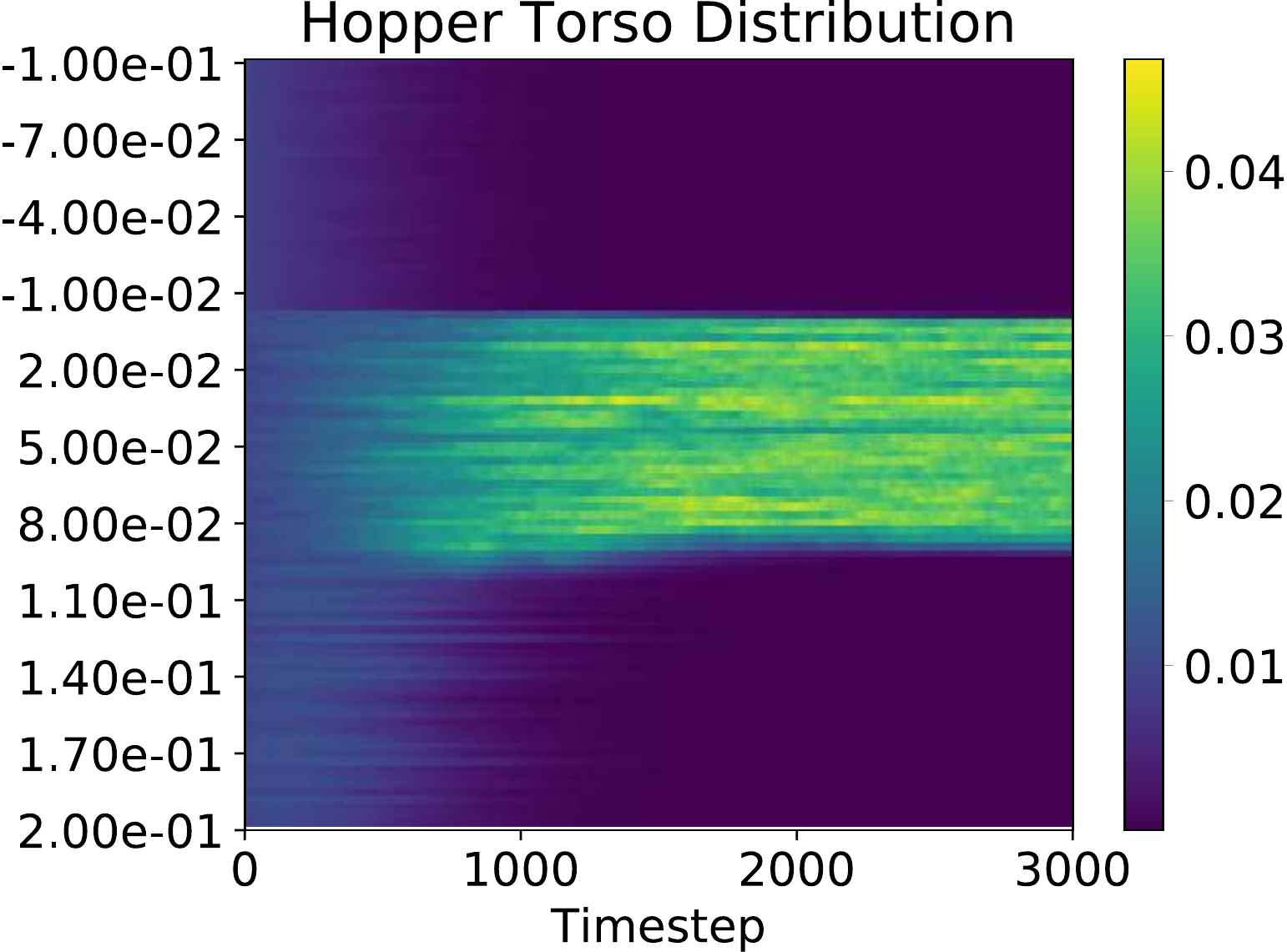}
        \label{fig:hop_torso_distr}
        \vspace{-1.5em}
        \caption{\small Torso size}
    \end{subfigure}
    ~
    \begin{subfigure}[b]{0.2\textwidth}
        \includegraphics[width=\textwidth]{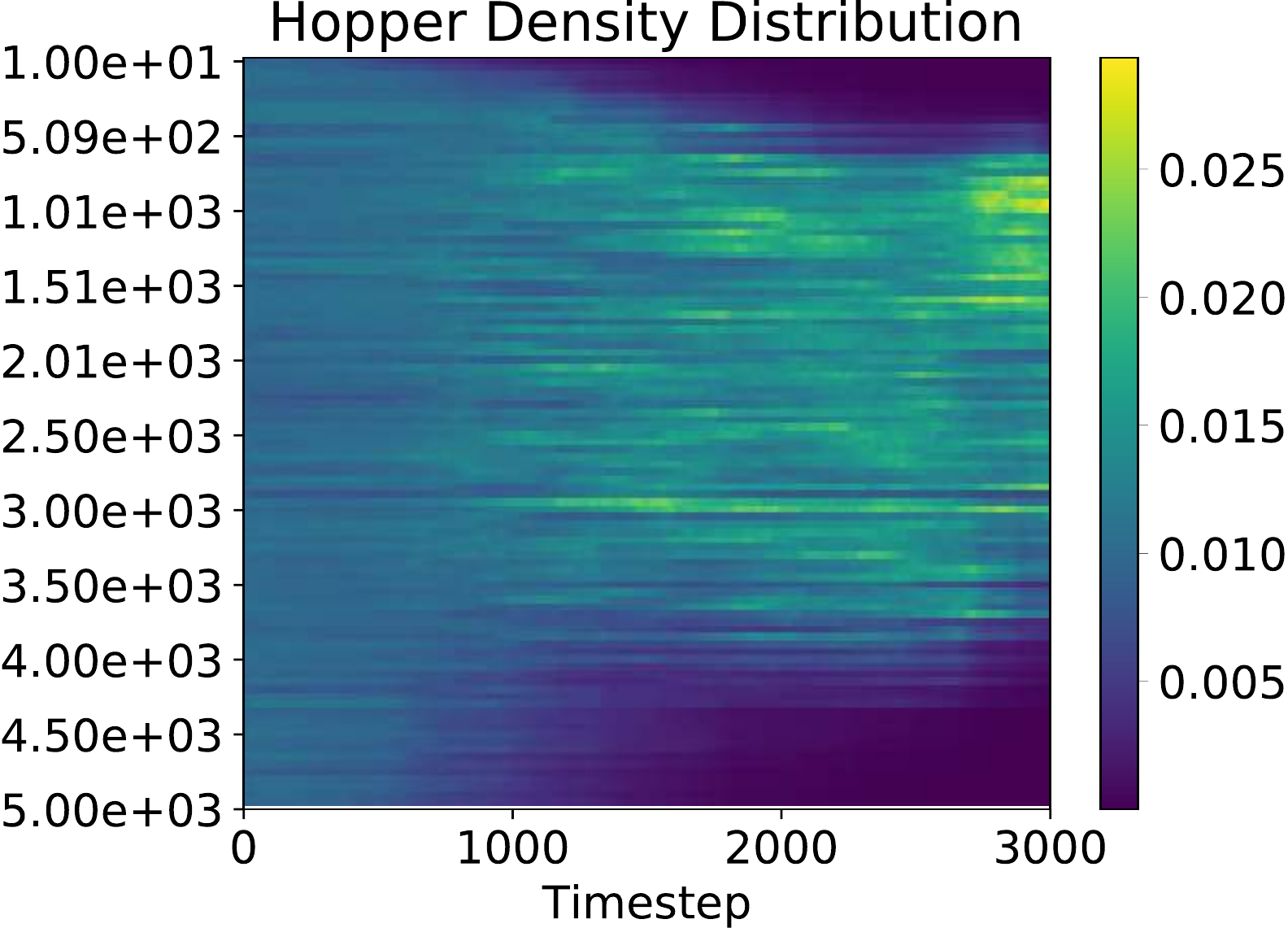}
        \label{fig:hop_density_distr}
        \vspace{-1.5em}
        \caption{\small Density}
    \end{subfigure}
    ~
    \begin{subfigure}[b]{0.2\textwidth}
        \includegraphics[width=\textwidth]{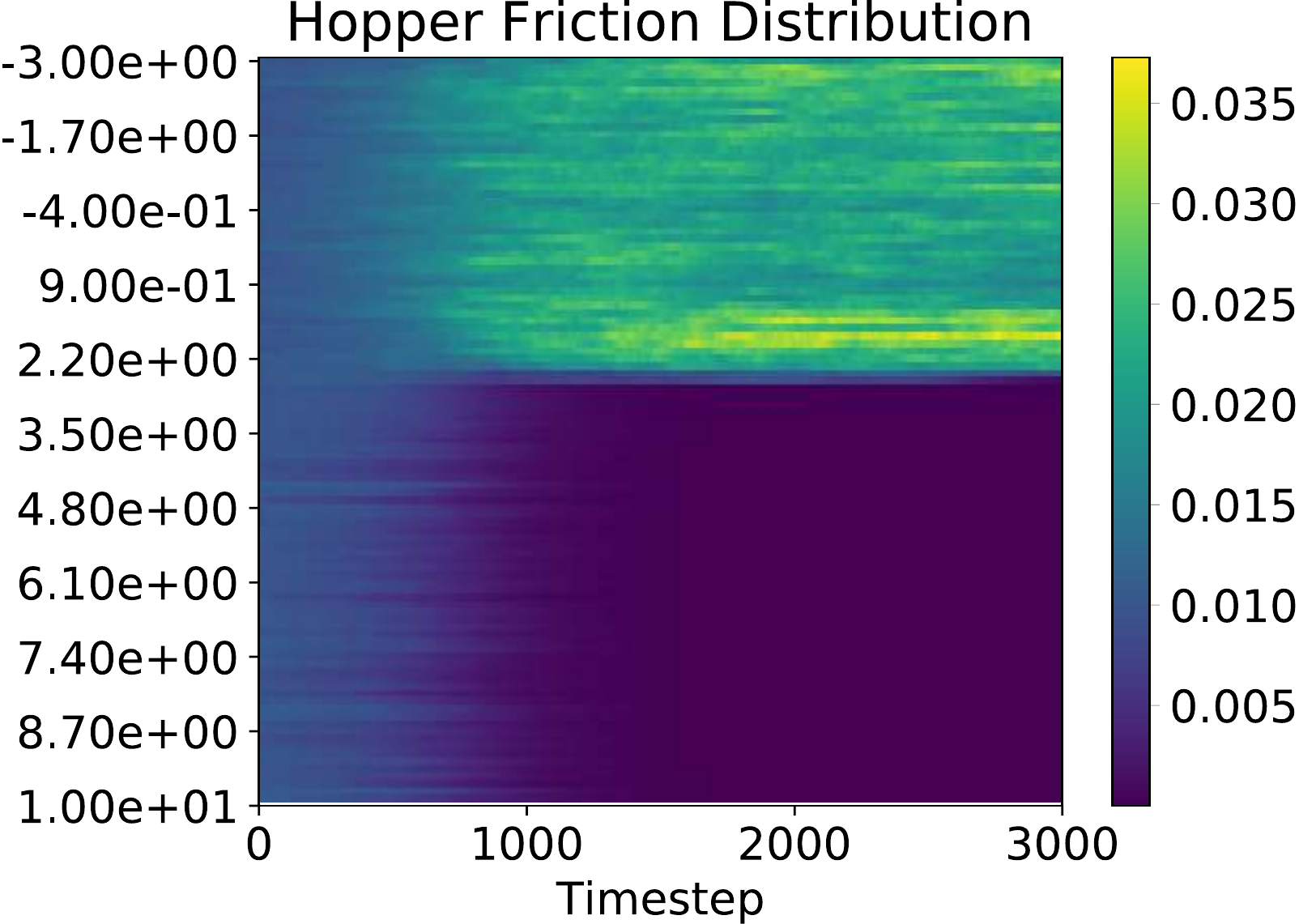}
        \label{fig:hop_friction_distr}
        \vspace{-1.5em}
        \caption{\small Friction}
    \end{subfigure}
    ~
    \begin{subfigure}[b]{0.2\textwidth}
        \includegraphics[width=\textwidth]{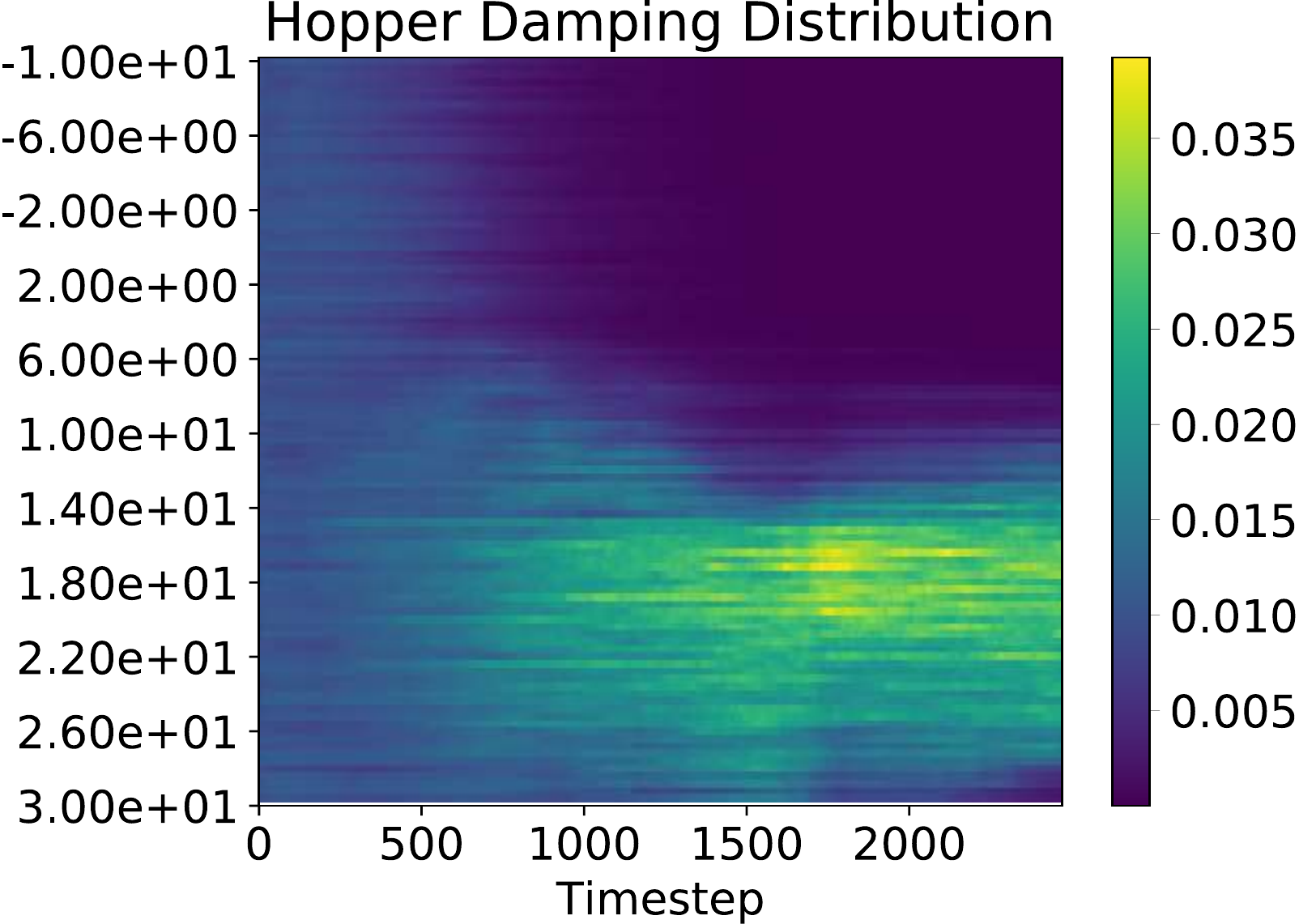}
        \label{fig:hop_damping_distr}
        \vspace{-1.5em}
        \caption{\small Damping}
    \end{subfigure}
    \caption{\small Evolution of the learned domain randomization distribution $p_{\phi}(z)$ over time for \textbf{Hopper}.  Each plot corresponds to a different experiments where we kept the other simulator parameters fixed at their default values. Lighter color corresponds to higher probabilities.}\label{fig:hop-dyna}
\end{figure}

\begin{figure}[t!]
    \centering
    \begin{subfigure}[b]{0.2\textwidth}
        \includegraphics[width=\textwidth]{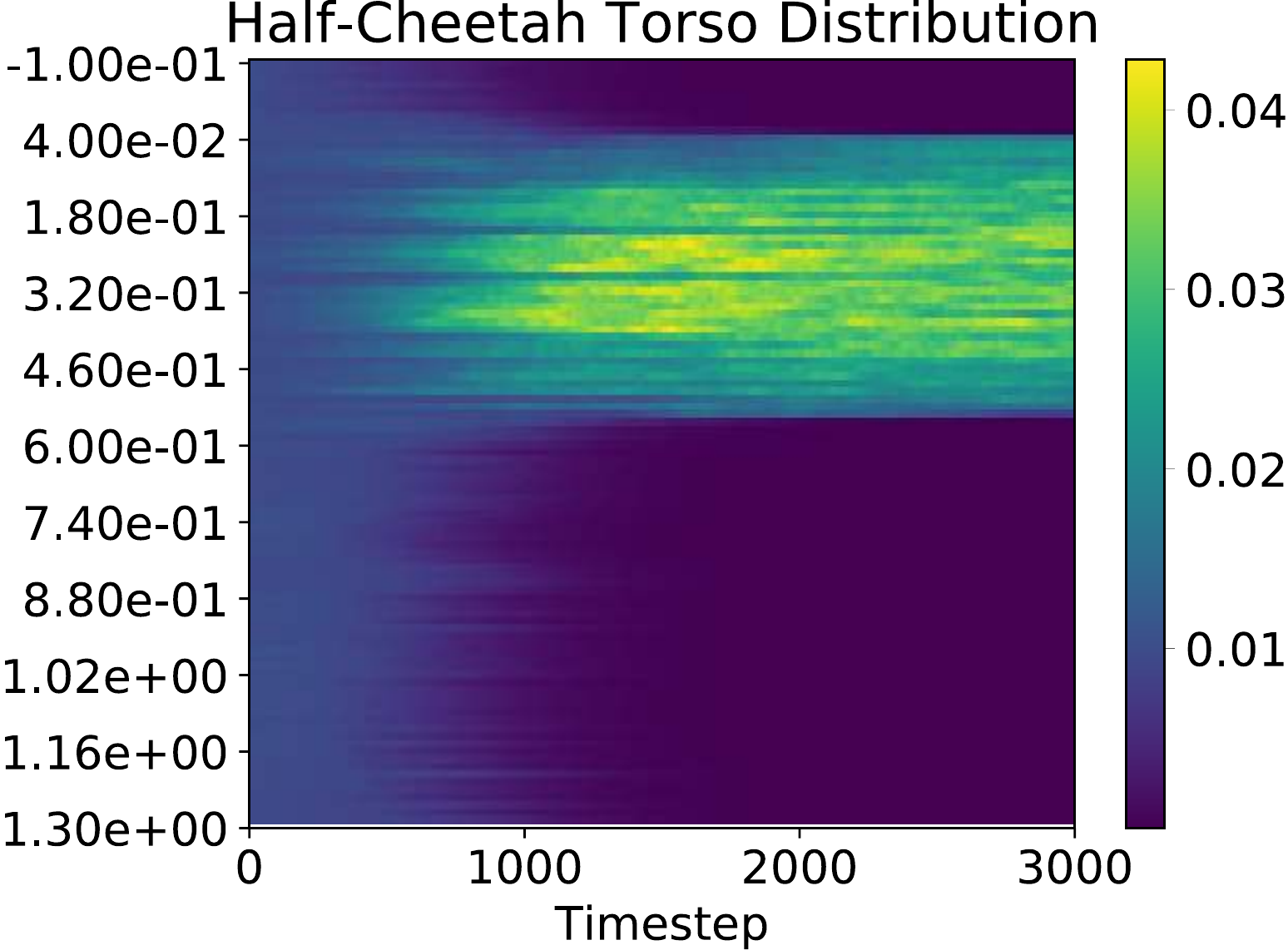}
        \label{fig:cheet_torso_distr}
        \vspace{-1.5em}
        \caption{\small Torso size}
    \end{subfigure}
    ~
    \begin{subfigure}[b]{0.2\textwidth}
        \includegraphics[width=\textwidth]{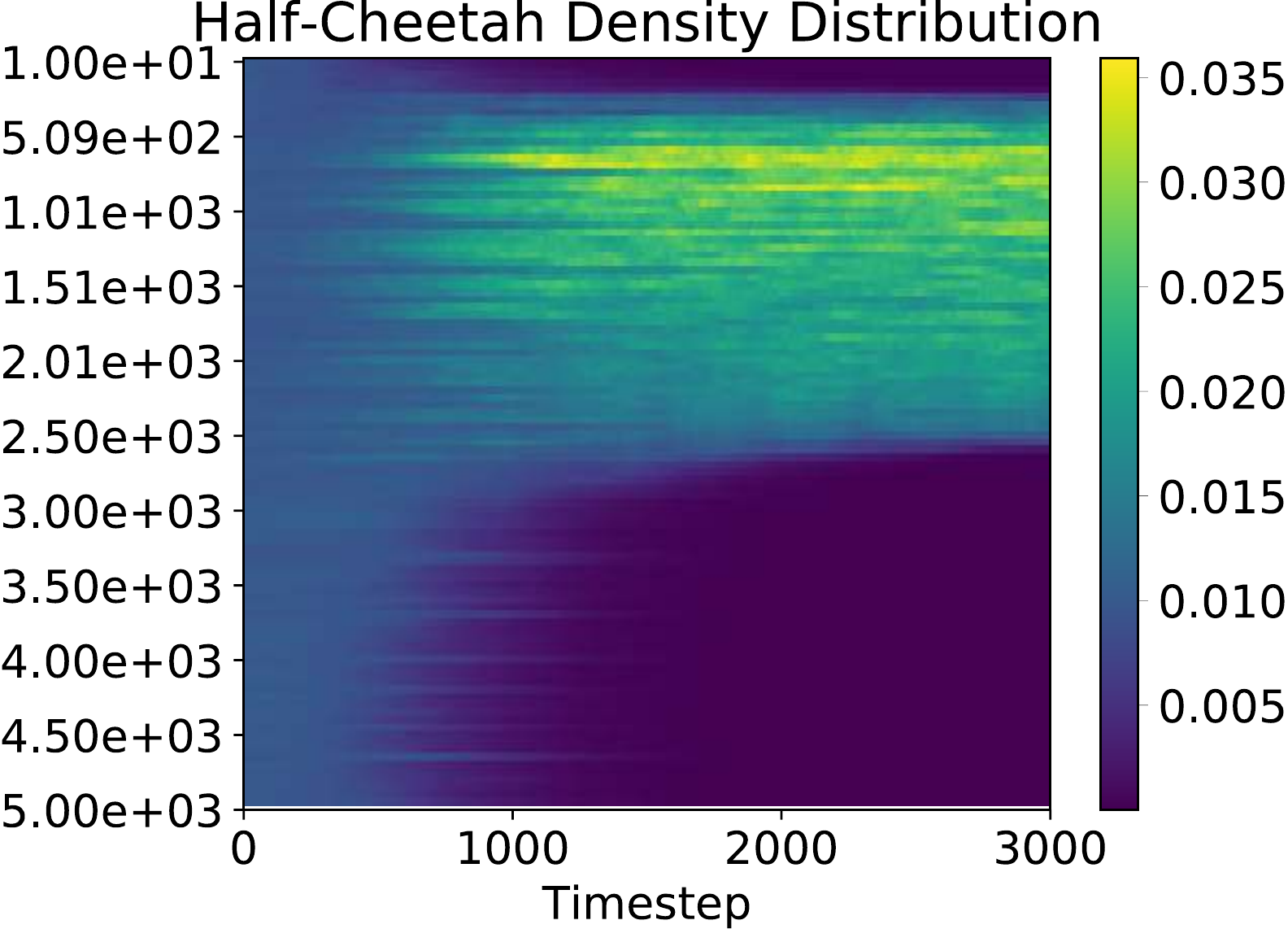}
        \label{fig:cheet_density_distr}
        \vspace{-1.5em}
        \caption{\small Density}
    \end{subfigure}
        ~
    \begin{subfigure}[b]{0.2\textwidth}
        \includegraphics[width=\textwidth]{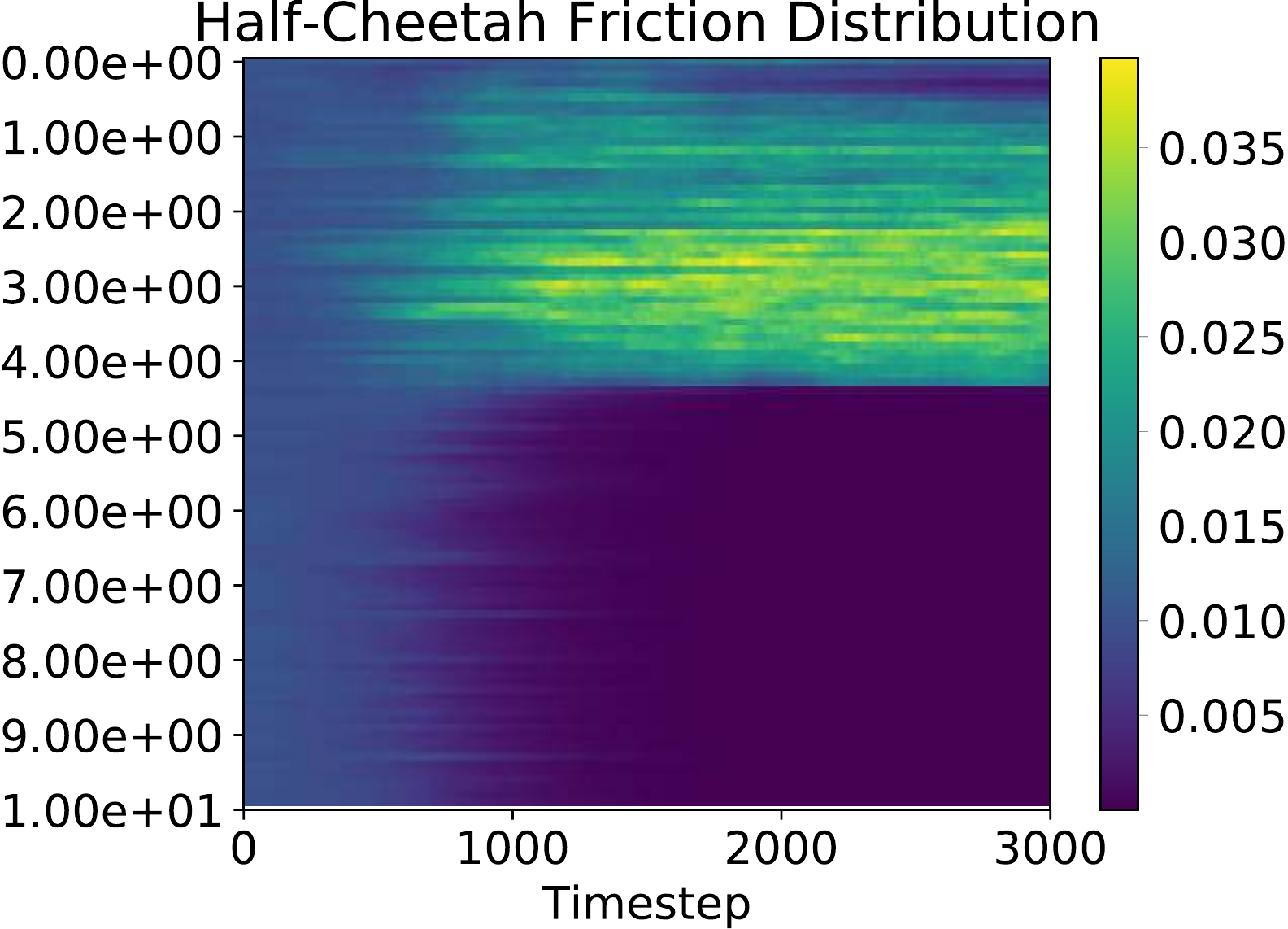}
        \label{fig:cheet_friction_distr}
        \vspace{-1.5em}
        \caption{\small Friction}
    \end{subfigure}
    ~
    \begin{subfigure}[b]{0.2\textwidth}
        \includegraphics[width=\textwidth]{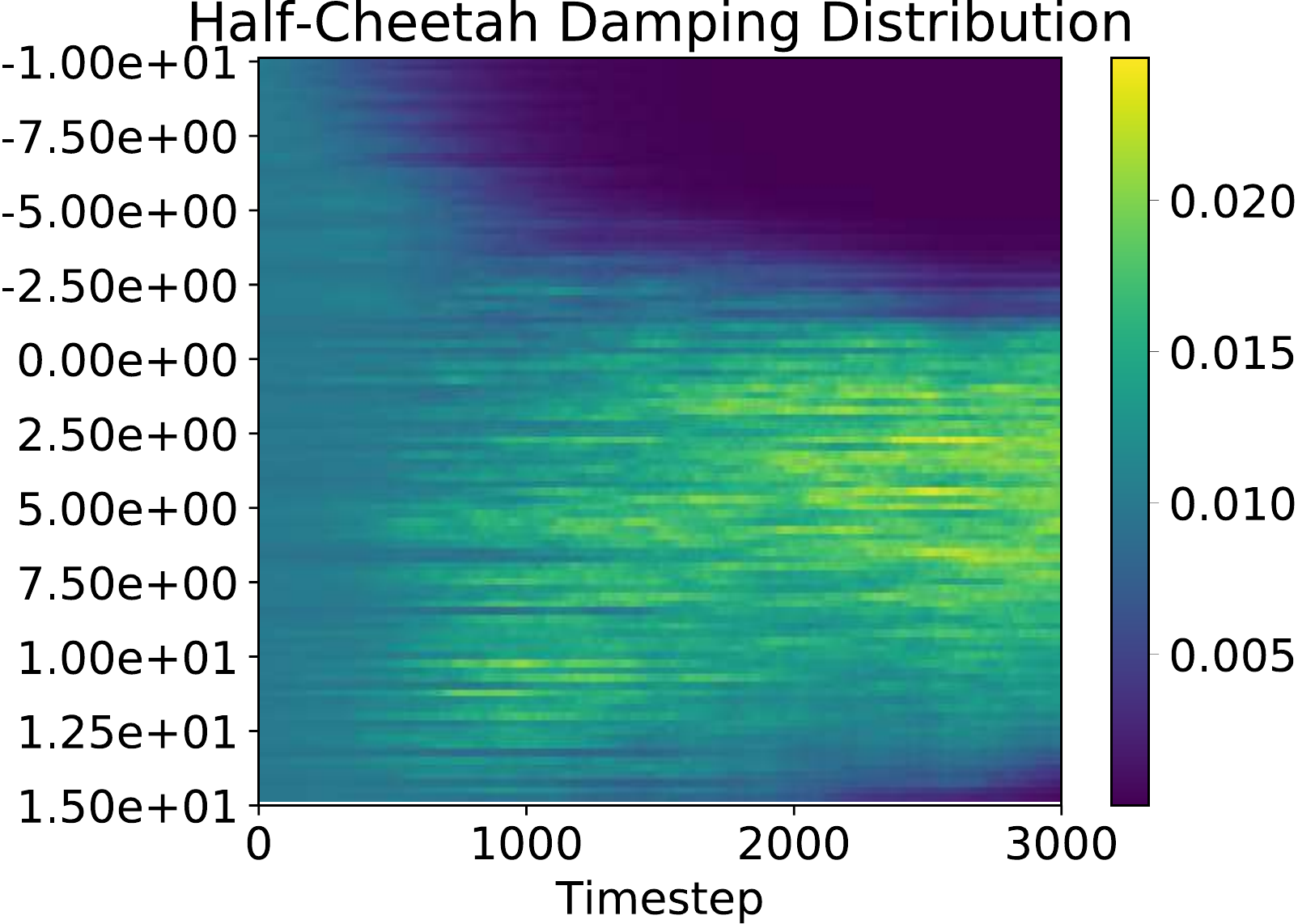}
        \label{fig:cheet_damping_distr}
        \vspace{-1.5em}
        \caption{\small Damping}
    \end{subfigure}
    \caption{\small Evolution of the learned domain randomization distribution $p_{\phi}(z)$  over time for \textbf{Half-Cheetah}. Note how the learned distribution assigns low probability to physically implausible environments (negative or large mass, large friction etc)}\label{fig:cheet-dyna}
\end{figure}

\begin{figure}[t!]
    \centering
    \begin{subfigure}[b]{0.14\textwidth}
        \includegraphics[width=\textwidth]{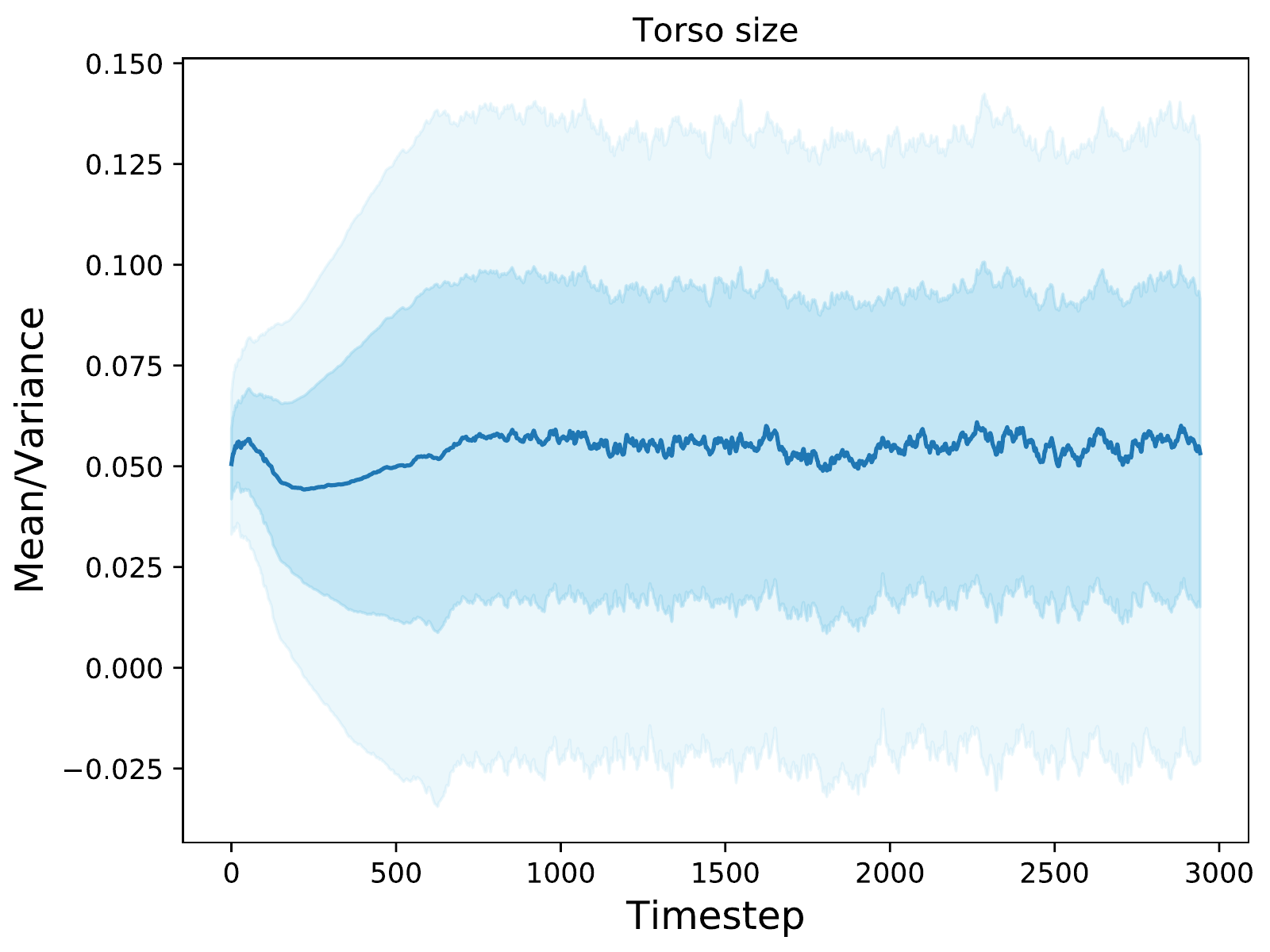}
        \label{fig:hop_multidim_torso_distr}
        \vspace{-1.5em}
        \caption{\small Torso size}
    \end{subfigure}
    ~
    \begin{subfigure}[b]{0.14\textwidth}
        \includegraphics[width=\textwidth]{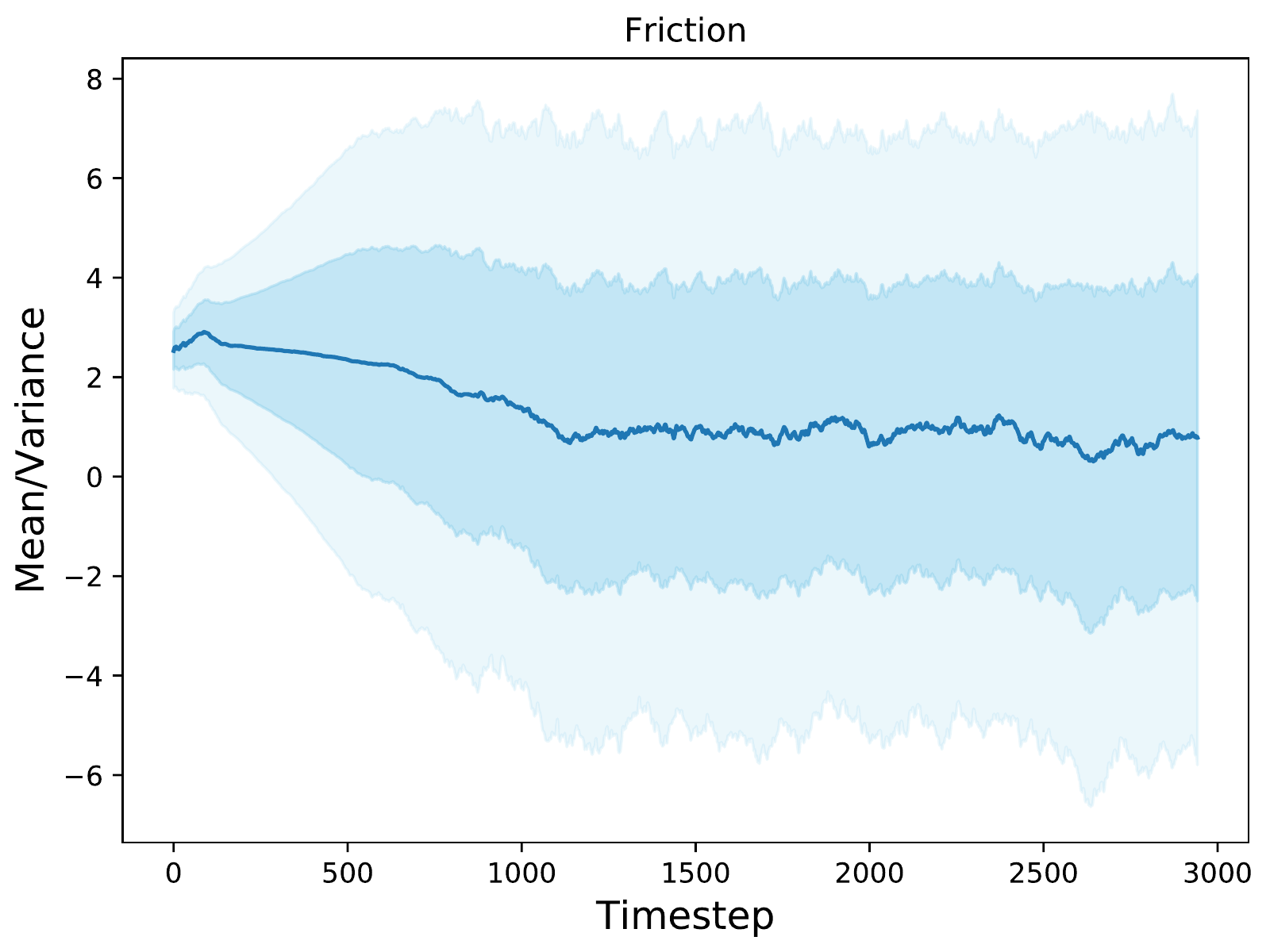}
        \label{fig:hop_multidim_friction_distr}
        \vspace{-1.5em}
        \caption{\small Friction}
    \end{subfigure}
    ~
    \begin{subfigure}[b]{0.14\textwidth}
        \includegraphics[width=\textwidth]{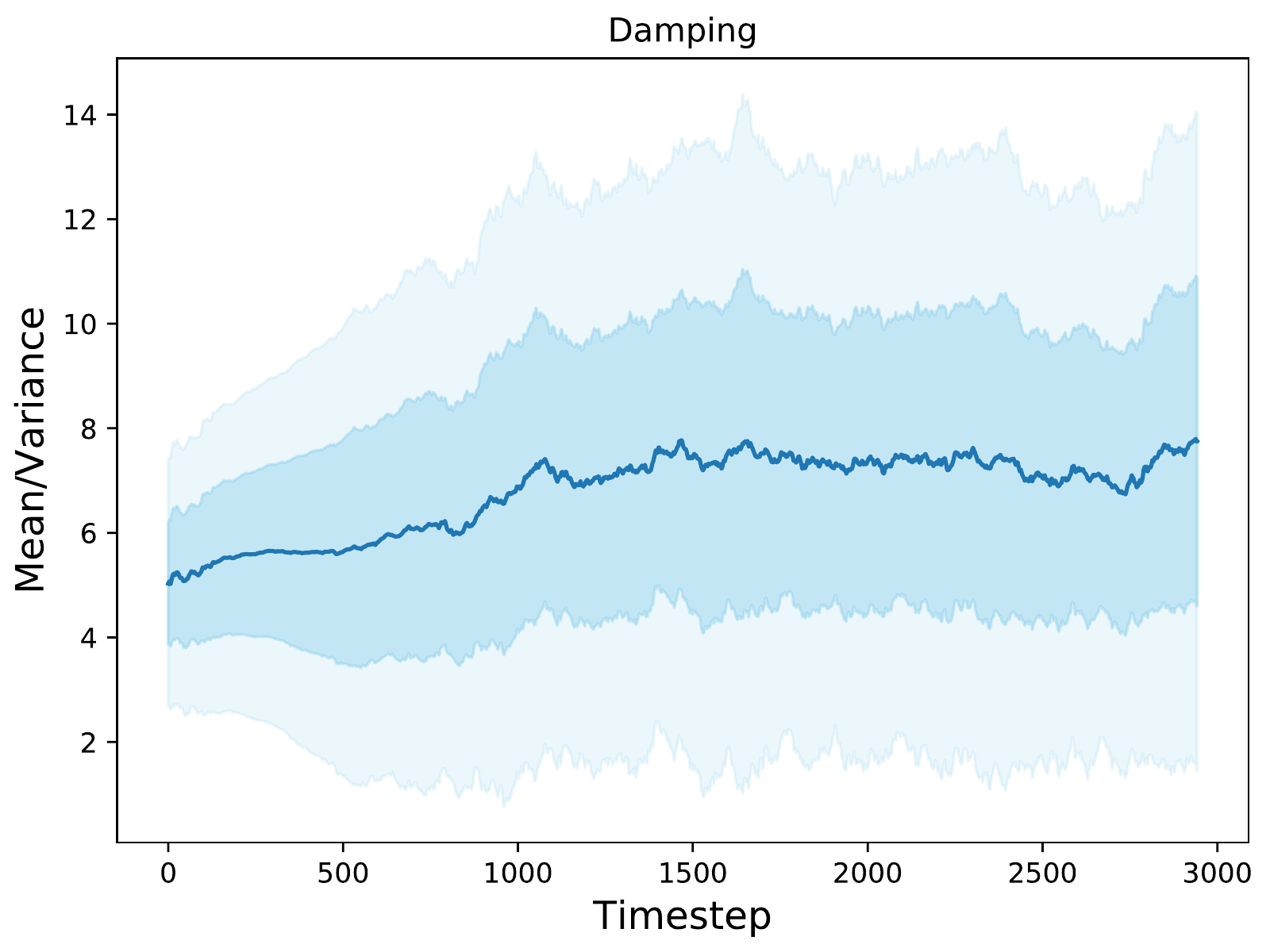}
        \label{fig:hop_multidim_damping_distr}
        \vspace{-1.5em}
        \caption{\small Damping}
    \end{subfigure}
    \caption{\small Evolution of the learned multidimensional domain randomization distribution $p_{\phi}(z)$ for \textbf{Hopper}.}
    \label{fig:hop-dyna-gaussian}
\end{figure}

\begin{figure}[t!]
    \centering
    \begin{subfigure}[b]{0.14\textwidth}
        \includegraphics[width=\textwidth]{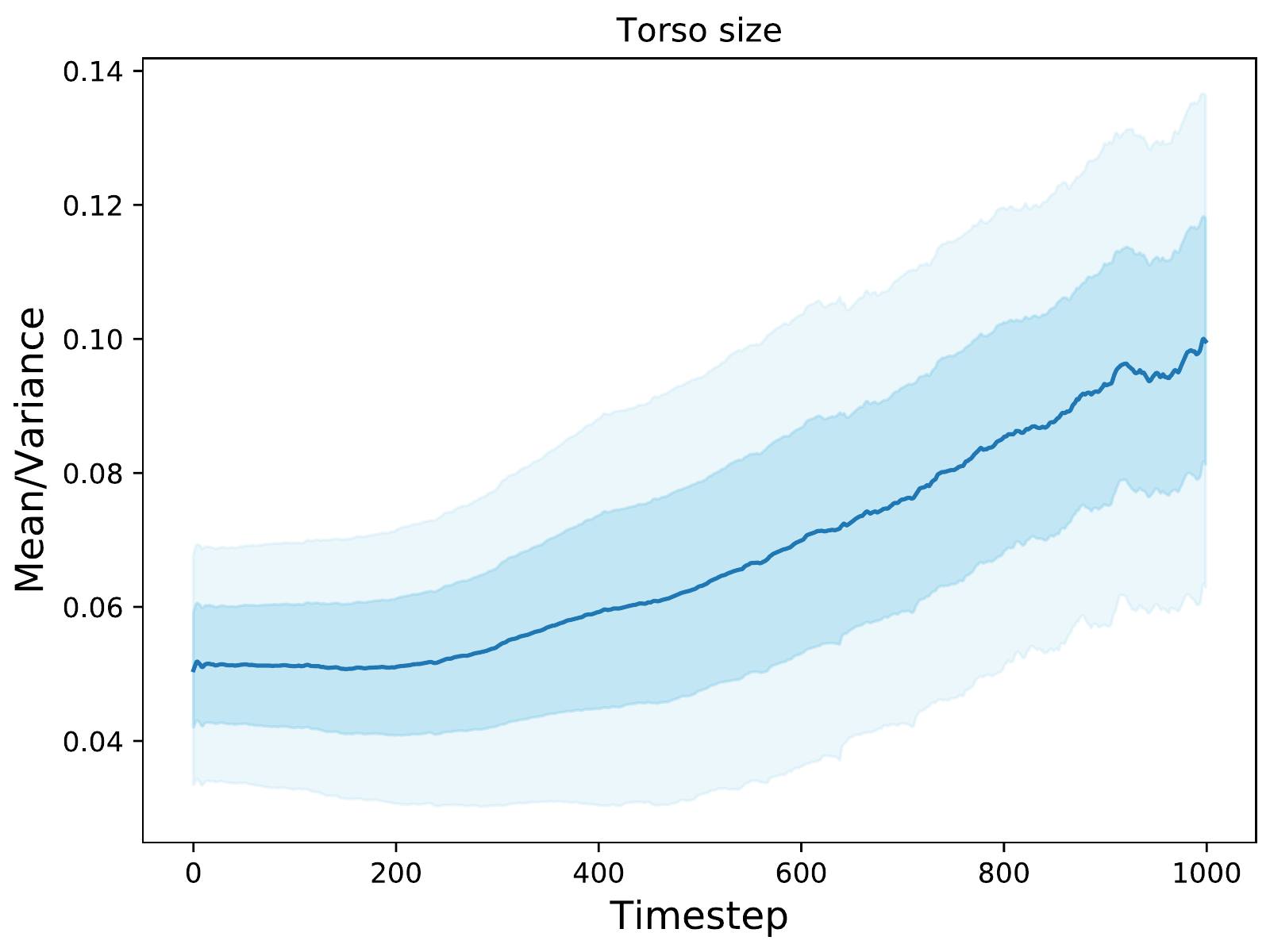}
        \label{fig:cheet_multi_torso_distr}
        \vspace{-1.5em}
        \caption{\small Torso size}
    \end{subfigure}
    ~
    \begin{subfigure}[b]{0.14\textwidth}
        \includegraphics[width=\textwidth]{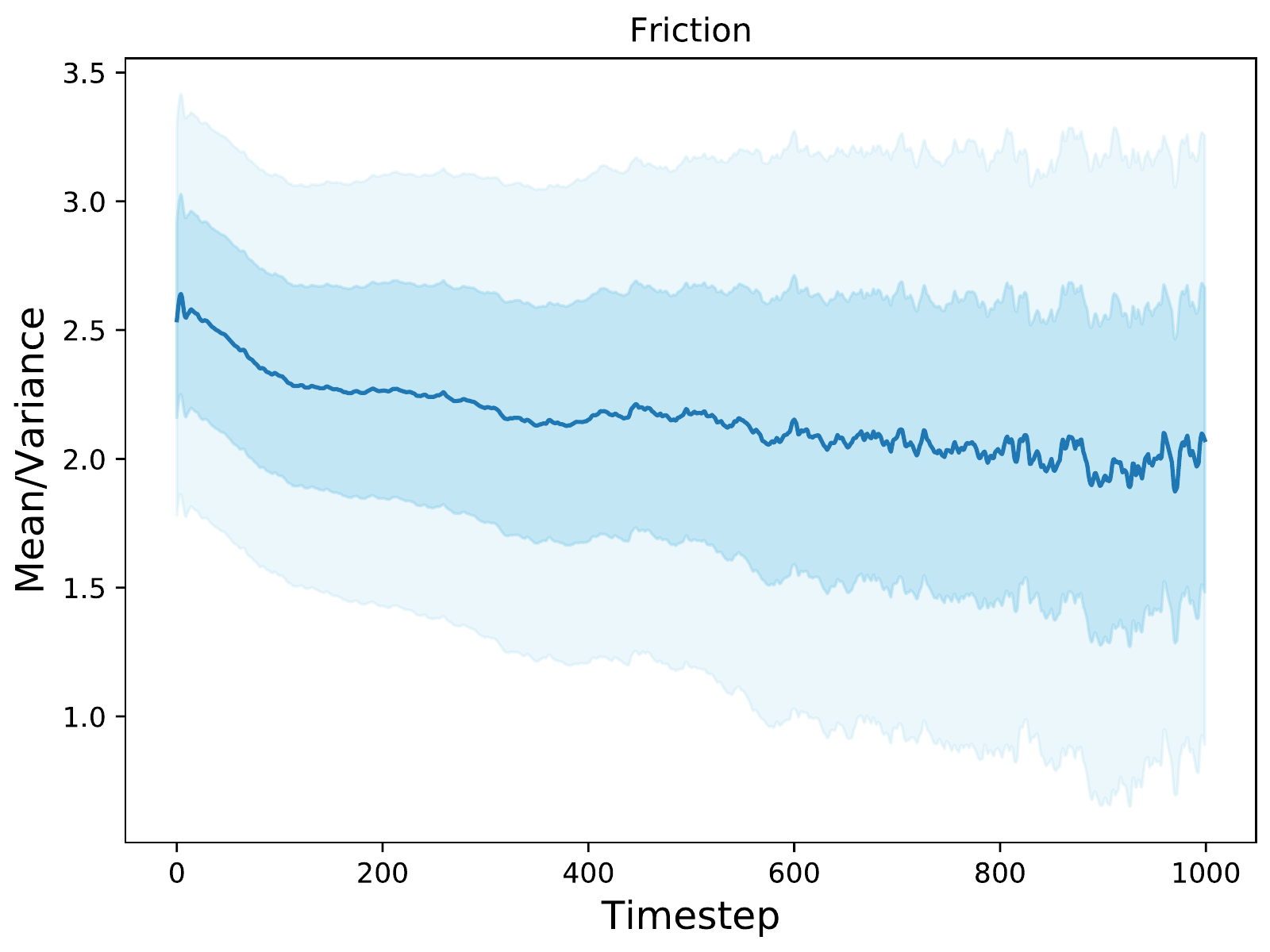}
        \label{fig:cheet_multi_friction_distr}
        \vspace{-1.5em}
        \caption{\small Friction}
    \end{subfigure}
    ~
    \begin{subfigure}[b]{0.14\textwidth}
        \includegraphics[width=\textwidth]{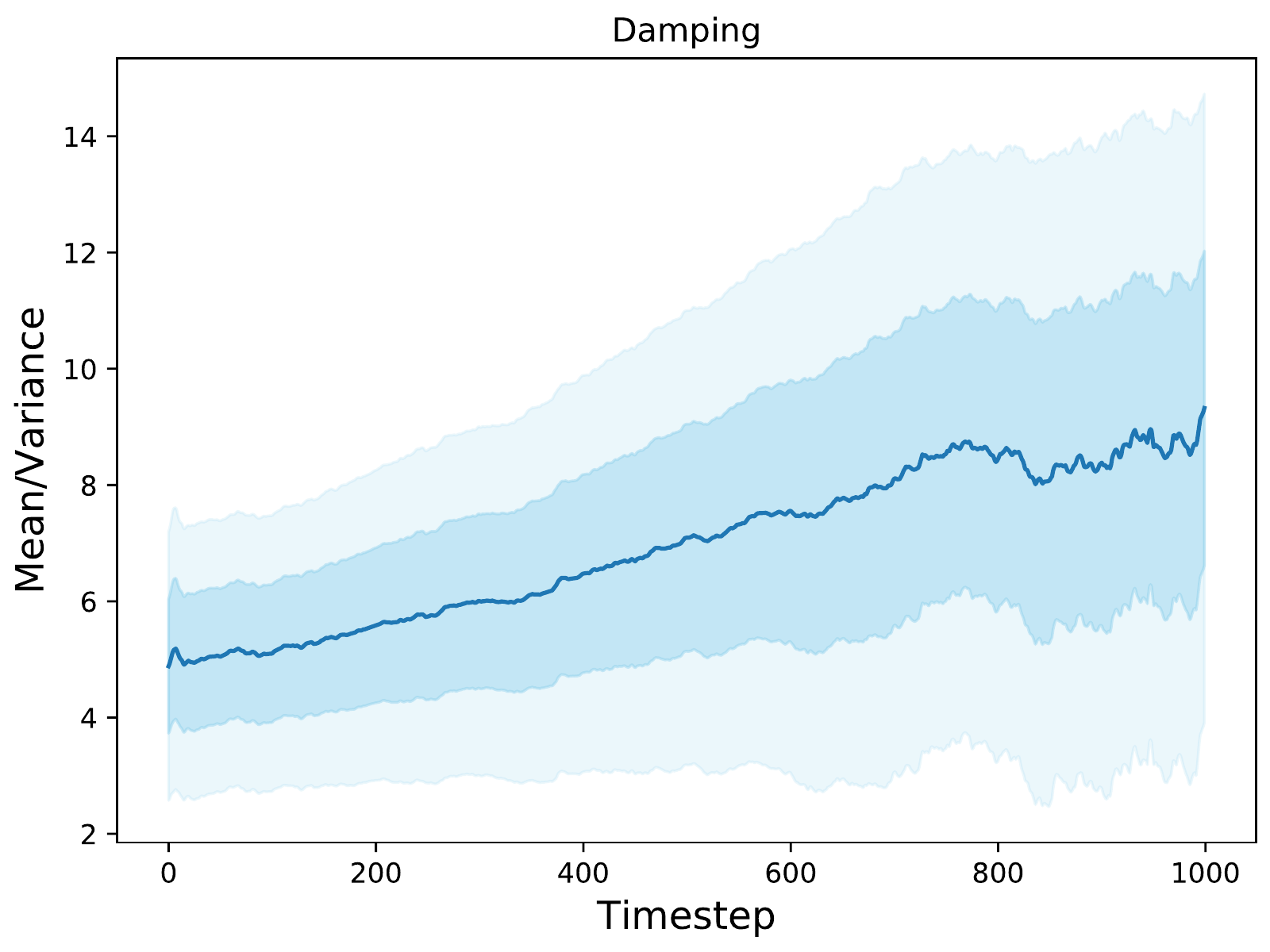}
        \label{fig:cheet_multi_damping_distr}
        \vspace{-1.5em}
        \caption{\small Damping}
    \end{subfigure}
    \caption{\small Evolution of the learned multidimensional domain randomization distribution $p_{\phi}(z)$ for \textbf{Half-Cheetah}.}
    \label{fig:cheet-dyna-gaussian}
\end{figure}

\noindent\textbf{Learned Distribution Ranges: }
Table~\ref{table:learned_ranges} shows the ranges for $p(z)$ and the final equivalent ranges for the distributions found by our method, for the single-dimensional discrete distribution learning experiments. 
Figures~\ref{fig:hop-dyna}, \ref{fig:cheet-dyna} \ref {fig:hop-dyna-gaussian} and \ref {fig:cheet-dyna-gaussian} show the evolution of $p_{\phi}(z)$ during training, using our method. In figures~\ref{fig:hop-dyna}, \ref{fig:cheet-dyna}, each plot corresponds to a separate domain randomization experiment, where we randomized one different simulator parameter while keeping the rest fixed. 
\begin{table*}[t!]
\centering
\resizebox{0.65\textwidth}{!}{
\begin{tabular}{c c c | c }
\multicolumn{3}{c}{} \\
\midrule
Environment & Parameters & Initial Train/Test Distribution & Converged Ranges \\
\midrule
\multirow{4}{*}{Hopper}    & Torso size    & $[-0.1, 0.2]$    &  $[0.0015, 0.09]$\\ 
                           & Density       & $[10, 5000]$     &  $[400, 4000]$\\
                           & Friction      & $[-3.0, 10.0]$   &  $[-3.0, 3.0]$\\
                           & Joint Damping & $[-10.0, 30.0]$  &  $[-1.0, 26]$\\
\bottomrule
\multirow{4}{*}{Half-Cheetah} & Torso size    & $[-0.1, 2.0]$    &  $[0.055, 0.52]$\\ 
                           & Density       & $[10, 5000]$     &  $[285, 2420]$\\
                           & Friction      & $[-3.0, 10.0]$   &  $[0.65, 4.39]$\\
                           & Joint Damping & $[-10.0, 30.0]$  &  $[-1.65, 12.9]$\\
\bottomrule   
\end{tabular}
}
\caption{\small Ranges of parameters for each environment, in the beginning of training and the equivalent ranges found by the algorithm, obtained by fitting an uniform distribution to the final learned distribution.}
\label{table:learned_ranges}
\end{table*}

Initially, each of these distribution is uniform. As the agent becomes better over the training distribution, it becomes easier to discriminate between promising environments (where the task is solvable) and impossible ones where rewards stay at low values. After around $1500$ epochs, the distributions have converged to their final distributions. For Hopper, the learned distributions corresponds closely with the environments where we can find a policy using vanilla policy gradient methods from scratch. To determine the consistency of these results, we ran the Hopper torso size experiment 7 times, and fitted the parameters of a uniform distribution to the resulting $p_{\phi}(z)$. The mean ranges ($\pm$ one standard deviation) across the 7 experiments were $[0.00086 \pm 0.00159, 0.09275 \pm 0.00342]$, which provides some evidence for the reproducibility of our method.
For figures~\ref{fig:hop-dyna}, \ref{fig:cheet-dyna}, all plots correspond to a multidimensional context experiment. Since the contexts are being learned jointly, the final distributions correspond to the joint distribution of all contexts. These plots correspond to the 2D projection of each context distribution.

\noindent\textbf{Learned vs Fixed Domain Randomization: }
We compare the jumpstart and asymptotic performance between learning the domain randomization distribution and keeping it fixed. Our results show our method, using PPO as the policy optimizer (\textbf{LSDR}) vs keeping the domain randomization distribution fixed (\textbf{Fixed-DR}) which corresponds to keeping the domain randomization distribution fixed. 
We ran the same experiments for \textit{Hopper} and \textit{Half-Cheetah}.

\begin{figure}[t!]
    \centering
    \begin{subfigure}[b]{0.4\textwidth}
        \includegraphics[width=\textwidth]{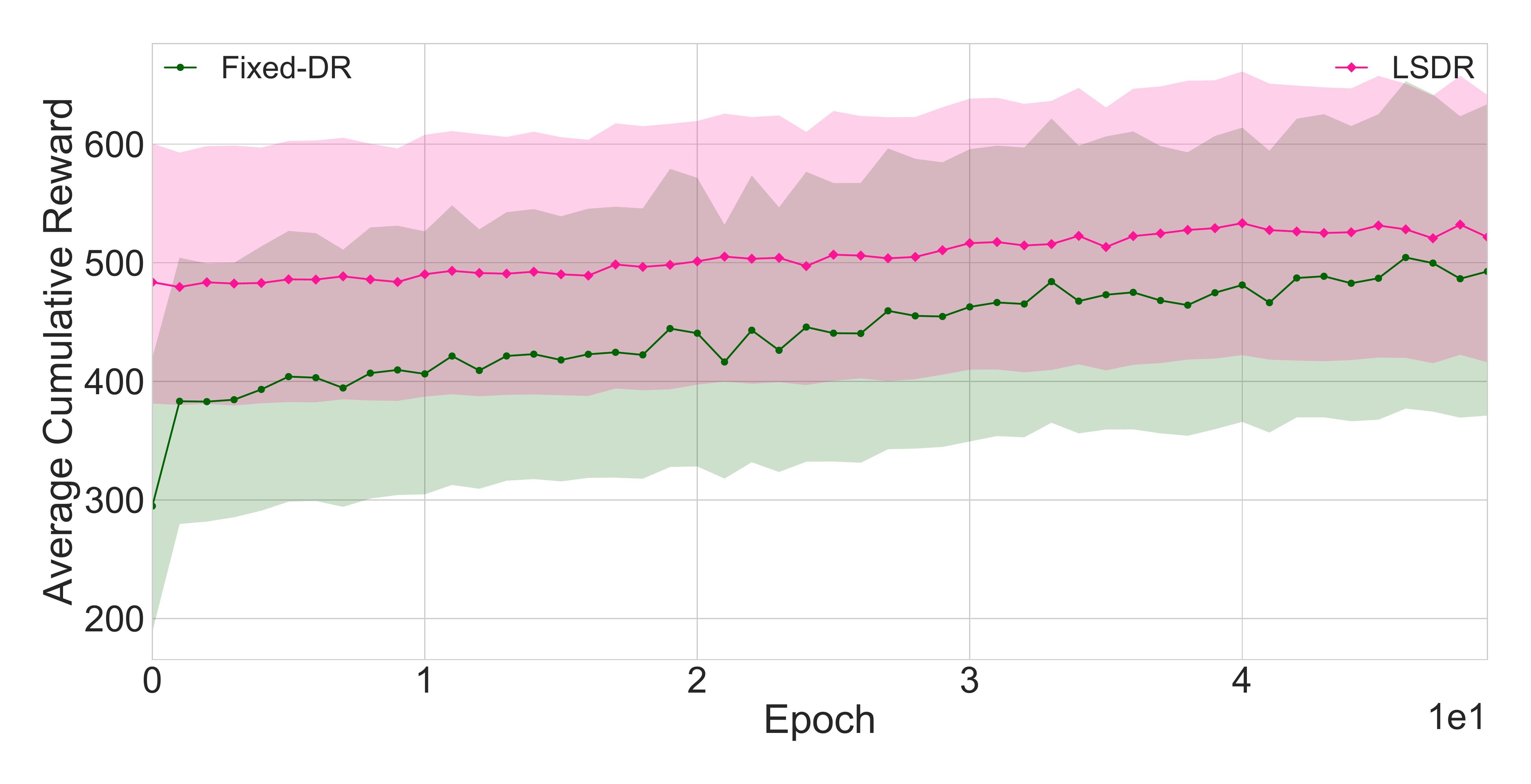}
        \label{fig:hopper-generalization}
        \vspace{-1em}
    \end{subfigure}
    ~ 
    \begin{subfigure}[b]{0.4\textwidth}
        \includegraphics[width=\textwidth]{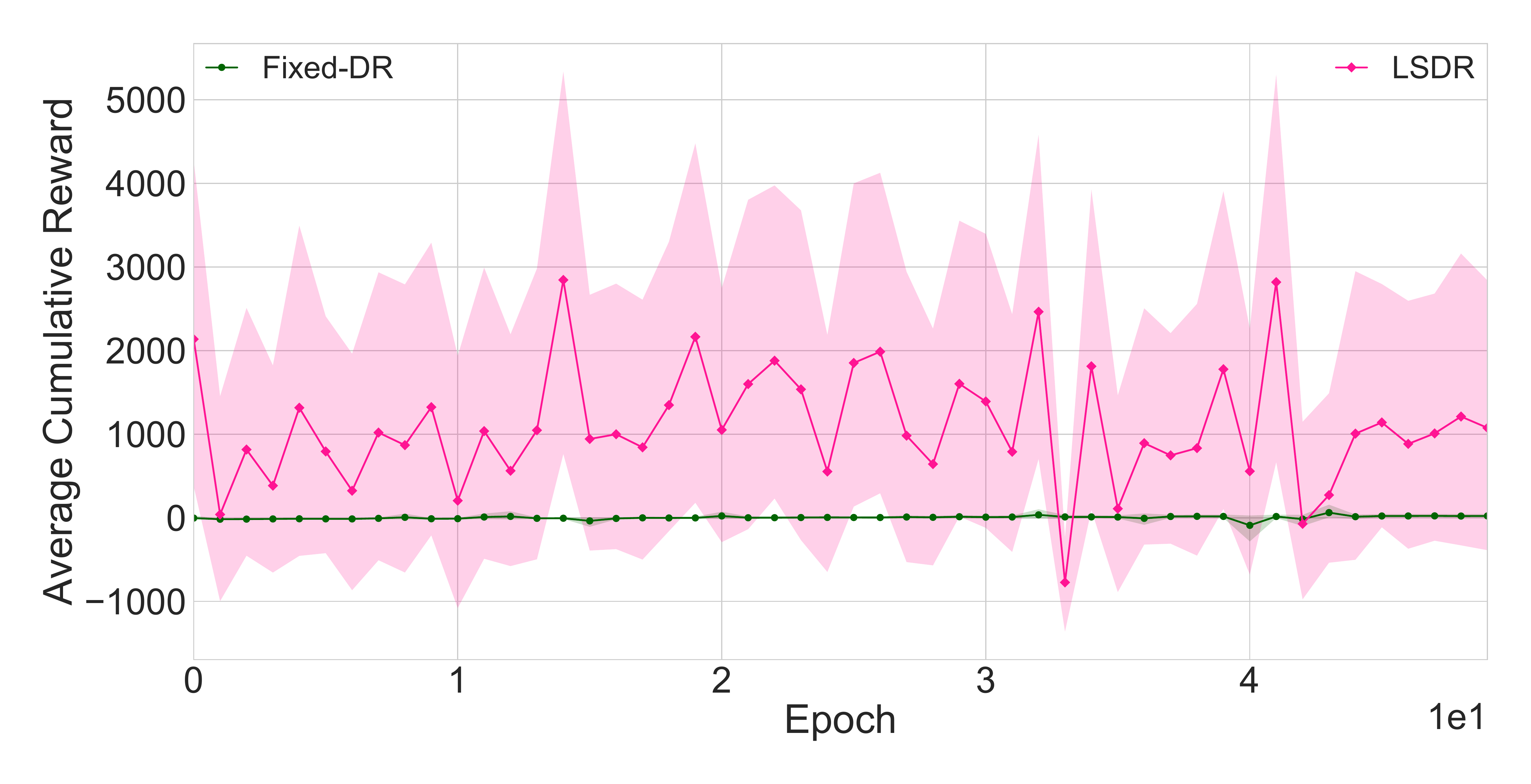}
        \label{fig:cheetah-generalization}
        \vspace{-1em}
    \end{subfigure}
    \caption{\small Comparison of test-time performance on Hopper (top) and Half-Cheetah (bottom) between fixed vs learned domain randomization in the policy inputs. Note that running experiments on the test range, will result in environments that may not necessarily be solvable and this results in additional variance on the performance across seeds.}
    \label{fig:hop-cheet-generalization}
\end{figure}

Figure~\ref{fig:hop-cheet-generalization} depict learning curves when fine tuning the policy at test-time, for torso size randomization. All the methods start with the same random seed at training time across $100$ seeds. The policies are trained for $3000$ epochs, where we collect $B=4000$ samples per epoch for the policy update. For the distribution update, we collect $K=10$ additional trajectories and run the gradient update for $M=10$ steps (without re-sampling new trajectories). 
To ensure fairness in terms of the sampled contexts, we report the comparison over the full test range (the support of $p(z)$). From these results, it is clear that learning the domain randomization distribution improves on the jump-start and asymptotic performance over using fixed domain randomization. In this case we test the policy generalization on the wide test distribution, which includes contexts that are not solvable; i.e. the optimal policy found by vanilla-PPO does not result in successful locomotion on the such extreme range\footnote{Successful policies on Hopper obtain cumulative rewards of at least $2500$. For Half-Cheetah, the rewards are greater than 0 when the robot successfully moves forward.}. For the multidimensional Hopper experiments, the results are shown in Figure~\ref{fig:hopper_grid_comparison}. In this case, we evaluated the performance of the policy at the last training iteration, with 10 rollouts per context. For the independently trained policies, we ran PPO for each context in the grid for $1000$ epochs, each with $4000$ steps of experience. While our method does not match the performance of the exhaustive grid training, it is clearly better than training on an uniform DR distribution. Furthermore, our method observes considerably less experience data ($1.2e7$ steps) than the exhaustive grid training ($4e9$ steps). We expect to get better performance by increasing the per-epoch experience, as done in~\cite{yu2017preparing}.


\noindent\textbf{Using a different policy optimizer}
\label{sec:epopt}
We also experimented with using EPOpt-PPO~\cite{rajeswaran2016epopt} as the policy optimizer in Algorithm~(\ref{alg:LSDR-policy}). The motivation for this is to mitigate the bias towards environments with higher cumulative rewards $J_{M_z}(\pi)$ early during training. EPOpt encourages improving the policy on the worst performing environments, at the expense of collecting more data per epoch. At each training epoch, we obtain $100$ samples from the training distribution and obtain trajectories by executing the policy once on each of the corresponding environments. From the resulting $100$ trajectories, we use the $10\%$ trajectories that resulted in the lowest rewards to fill the buffer for a PPO policy update, discarding the rest of the trajectories.

\begin{figure}[t!]
    \centering
    \includegraphics[width=0.4\textwidth]{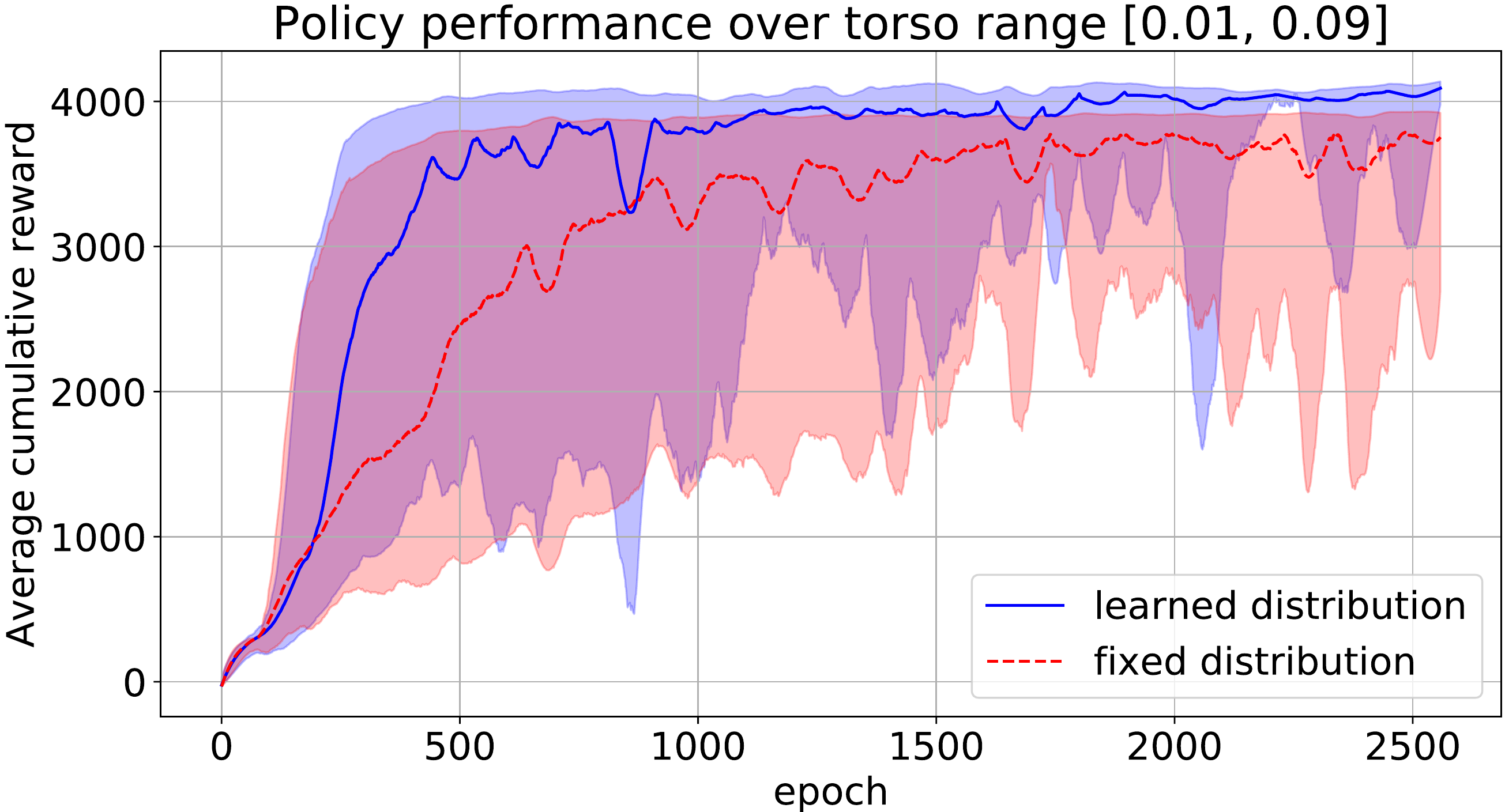}
    \caption{\small Learning curves for torso DR on Hopper using EPOpt. Lines represent mean performance, while the shaded regions correspond to the maximum and minimum performance (smoothed with a 5th order Savitzky-Golay filter and a window of 10 epochs)}
    \label{fig:epopt_minmax}
\end{figure}


Figure~\ref{fig:epopt_minmax} compares the effect of learning the domain randomization distribution vs using a fixed wide range in this setting. We found that learning the domain randomization distribution resulted in faster convergence to high reward policies over the evaluation range [0.01, 0.09]: our method attains cumulative rewards greater than 3000 in about 1/2 the epochs required with uniform DR. Our method also results in better asymptotic performance: our method finds policies with better and more consistent performance (less variance) as seen in Figure~\ref{fig:epopt_minmax}. We believe this could be a consequence of lower variance in the policy gradient estimates, as the the learned $p_{\phi}(z)$ has lower variance than $p(z)$. Interestingly, using EPOpt resulted in a distribution with a wider torso size range than vanilla PPO, from approximately $0.0$ to $0.14$, demonstrating that optimizing worst case performance does help in alleviating the bias towards high reward environments.

\section{Discussion}
By allowing the agent to learn a good representative distribution, we are able to learn to solve difficult control tasks that heavily rely on a good initial domain randomization range.
Our main experimental validation of domain randomization distribution learning is in the domain of simulated robotic locomotion.
As shown in our experiments, our method is not sensitive to the initial domain randomization distribution and is able to converge to a more diverse range, while staying within the feasible range. We experimented with single dimensional discrete and multivariate Gaussian distributions for learning the DR distributions. Our results show the benefit of learning a domain randomization distribution when compared to the previously accepted practice: keeping it fixed. Further extension of this work will explore better objectives for learning the distribution and more flexible parameterizations for the DR distribution.


Unscaled rewards, non-linear dynamics and many other complex factors make the manual setting of DR ranges ill-suited for human intuition. Automating this step is crucial to improving the robustness of simulation-to-real transfer and solving today's wide range of robotics challenges; our method is a new step in this fruitful direction.

\bibliography{root.bib}{}

\begin{thebibliography}{10}
\providecommand{\url}[1]{#1}
\csname url@samestyle\endcsname
\providecommand{\newblock}{\relax}
\providecommand{\bibinfo}[2]{#2}
\providecommand{\BIBentrySTDinterwordspacing}{\spaceskip=0pt\relax}
\providecommand{\BIBentryALTinterwordstretchfactor}{4}
\providecommand{\BIBentryALTinterwordspacing}{\spaceskip=\fontdimen2\font plus
\BIBentryALTinterwordstretchfactor\fontdimen3\font minus
  \fontdimen4\font\relax}
\providecommand{\BIBforeignlanguage}[2]{{%
\expandafter\ifx\csname l@#1\endcsname\relax
\typeout{** WARNING: IEEEtran.bst: No hyphenation pattern has been}%
\typeout{** loaded for the language `#1'. Using the pattern for}%
\typeout{** the default language instead.}%
\else
\language=\csname l@#1\endcsname
\fi
#2}}
\providecommand{\BIBdecl}{\relax}
\BIBdecl

\bibitem{jakobi1995noise}
N.~Jakobi, P.~Husbands, and I.~Harvey, ``Noise and the reality gap: The use of
  simulation in evolutionary robotics,'' in \emph{European Conference on
  Artificial Life}.\hskip 1em plus 0.5em minus 0.4em\relax Springer, 1995, pp.
  704--720.

\bibitem{peng2018sim}
X.~B. Peng, M.~Andrychowicz, W.~Zaremba, and P.~Abbeel, ``Sim-to-real transfer
  of robotic control with dynamics randomization,'' in \emph{2018 IEEE
  International Conference on Robotics and Automation (ICRA)}.\hskip 1em plus
  0.5em minus 0.4em\relax IEEE, 2018, pp. 1--8.

\bibitem{openai2018learning}
M.~Andrychowicz, B.~Baker, M.~Chociej, R.~Jozefowicz, B.~McGrew, J.~Pachocki,
  A.~Petron, M.~Plappert, G.~Powell, A.~Ray \emph{et~al.}, ``Learning dexterous
  in-hand manipulation,'' \emph{arXiv preprint arXiv:1808.00177}, 2018.

\bibitem{chen2018hardware}
T.~Chen, A.~Murali, and A.~Gupta, ``Hardware conditioned policies for
  multi-robot transfer learning,'' in \emph{Advances in Neural Information
  Processing Systems}, 2018, pp. 9355--9366.

\bibitem{rajeswaran2016epopt}
\BIBentryALTinterwordspacing
A.~Rajeswaran, S.~Ghotra, S.~Levine, and B.~Ravindran, ``{EPOpt}: Learning
  robust neural network policies using model ensembles,'' \emph{CoRR}, vol.
  abs/1610.01283, 2016. [Online]. Available:
  \url{http://arxiv.org/abs/1610.01283}
\BIBentrySTDinterwordspacing

\bibitem{packer2018assessing}
C.~Packer, K.~Gao, J.~Kos, P.~Kr{\"a}henb{\"u}hl, V.~Koltun, and D.~Song,
  ``Assessing generalization in deep reinforcement learning,'' \emph{arXiv
  preprint arXiv:1810.12282}, 2018.

\bibitem{chebotar2018closing}
Y.~Chebotar, A.~Handa, V.~Makoviychuk, M.~Macklin, J.~Issac, N.~Ratliff, and
  D.~Fox, ``Closing the sim-to-real loop: Adapting simulation randomization
  with real world experience,'' \emph{arXiv preprint arXiv:1810.05687}, 2018.

\bibitem{ramos2019bayessim}
F.~Ramos, R.~Possas, and D.~Fox, ``Bayessim: Adaptive domain randomization via
  probabilistic inference for robotics simulators,'' in \emph{Proceedings of
  Robotics: Science and Systems}, FreiburgimBreisgau, Germany, June 2019.

\bibitem{gym2016}
G.~Brockman, V.~Cheung, L.~Pettersson, J.~Schneider, J.~Schulman, J.~Tang, and
  W.~Zaremba, ``Openai gym,'' 2016.

\bibitem{Zames1981}
G.~{Zames}, ``Feedback and optimal sensitivity: Model reference
  transformations, multiplicative seminorms, and approximate inverses,''
  \emph{IEEE Transactions on Automatic Control}, vol.~26, no.~2, pp. 301--320,
  April 1981.

\bibitem{zhou1998robustControl}
K.~Zhou and J.~C. Doyle, \emph{Essentials of robust control}.\hskip 1em plus
  0.5em minus 0.4em\relax Prentice hall Upper Saddle River, NJ, 1998, vol. 104.

\bibitem{Caruana93multitasklearning}
R.~Caruana, ``Multitask learning: A knowledge-based source of inductive bias,''
  in \emph{Proceedings of the Tenth International Conference on Machine
  Learning}.\hskip 1em plus 0.5em minus 0.4em\relax Morgan Kaufmann, 1993, pp.
  41--48.

\bibitem{Baxter1997}
\BIBentryALTinterwordspacing
J.~Baxter, ``A bayesian/information theoretic model of learning to learn via
  multiple task sampling,'' \emph{Machine Learning}, vol.~28, no.~1, pp. 7--39,
  Jul 1997. [Online]. Available: \url{https://doi.org/10.1023/A:1007327622663}
\BIBentrySTDinterwordspacing

\bibitem{tamar2015optimizing}
A.~Tamar, Y.~Glassner, and S.~Mannor, ``Optimizing the cvar via sampling,'' in
  \emph{Twenty-Ninth AAAI Conference on Artificial Intelligence}, 2015.

\bibitem{fpo_opt}
\BIBentryALTinterwordspacing
S.~Paul, M.~A. Osborne, and S.~Whiteson, ``Contextual policy optimisation,''
  \emph{CoRR}, vol. abs/1805.10662, 2018. [Online]. Available:
  \url{http://arxiv.org/abs/1805.10662}
\BIBentrySTDinterwordspacing

\bibitem{mehta2019adr}
\BIBentryALTinterwordspacing
B.~Mehta, M.~Diaz, F.~Golemo, C.~J. Pal, and L.~Paull, ``Active domain
  randomization,'' \emph{CoRR}, vol. abs/1904.04762, 2019. [Online]. Available:
  \url{http://arxiv.org/abs/1904.04762}
\BIBentrySTDinterwordspacing

\bibitem{strategy_optim}
\BIBentryALTinterwordspacing
W.~Yu, C.~K. Liu, and G.~Turk, ``Policy transfer with strategy optimization,''
  \emph{CoRR}, vol. abs/1810.05751, 2018. [Online]. Available:
  \url{http://arxiv.org/abs/1810.05751}
\BIBentrySTDinterwordspacing

\bibitem{rakelly2019efficient}
K.~Rakelly, A.~Zhou, D.~Quillen, C.~Finn, and S.~Levine, ``Efficient off-policy
  meta-reinforcement learning via probabilistic context variables,''
  \emph{arXiv preprint arXiv:1903.08254}, 2019.

\bibitem{yu2017preparing}
W.~Yu, J.~Tan, C.~K. Liu, and G.~Turk, ``Preparing for the unknown: Learning a
  universal policy with online system identification,'' \emph{arXiv preprint
  arXiv:1702.02453}, 2017.

\bibitem{sutton2018reinforcement}
\BIBentryALTinterwordspacing
R.~Sutton and A.~Barto, \emph{Reinforcement Learning: An Introduction}, ser.
  Adaptive Computation and Machine Learning series.\hskip 1em plus 0.5em minus
  0.4em\relax MIT Press, 2018. [Online]. Available:
  \url{https://books.google.ca/books?id=6DKPtQEACAAJ}
\BIBentrySTDinterwordspacing

\bibitem{schulman2017proximal}
J.~Schulman, F.~Wolski, P.~Dhariwal, A.~Radford, and O.~Klimov, ``Proximal
  policy optimization algorithms,'' \emph{arXiv preprint arXiv:1707.06347},
  2017.

\bibitem{fu2006gradient}
M.~C. Fu, ``Gradient estimation,'' \emph{Handbooks in operations research and
  management science}, vol.~13, pp. 575--616, 2006.

\end{thebibliography}
\bibliographystyle{IEEEtran}

\end{document}